\documentclass[lettersize,journal]{IEEEtran}
\usepackage{amsmath,amsfonts}
\usepackage{algorithmic}
\usepackage{algorithm}
\usepackage{array}
\usepackage[caption=false,font=normalsize,labelfont=sf,textfont=sf]{subfig}
\usepackage{textcomp}
\usepackage{stfloats}
\usepackage{url}
\usepackage{verbatim}
\usepackage{graphicx}
\usepackage{cite}
\usepackage{graphicx}
\usepackage{amsmath}
\usepackage{amssymb}

\usepackage{xcolor}
\usepackage{enumitem}
\usepackage{xspace}
\usepackage{booktabs}
\usepackage{colortbl}
\usepackage{multirow}
\usepackage{makecell}
\usepackage{lipsum} %
\usepackage{gensymb} %
\usepackage{hyperref}
\usepackage{amsmath}
\usepackage{amssymb}
\usepackage{bbm} 

\newcommand{\Tref}[1]{Table~\ref{#1}}
\newcommand{\eref}[1]{Eq.~\eqref{#1}}

\newcommand{\fref}[1]{Fig.~\ref{#1}}
\newcommand{\Fref}[1]{Figure~\ref{#1}}
\newcommand{\sref}[1]{Sec.~\ref{#1}}

\usepackage{algorithm}
\usepackage{algorithmic}
\newcommand{\eg}{\emph{e.g.}}  %

\definecolor{MyDarkRed}{rgb}{0.66, 0.16, 0.16}
\definecolor{MyDarkBlue}{rgb}{0.16, 0.16, 0.66}

\newcommand{\zyqm}[1]{\textcolor{black}{#1}}

\newcommand{\MethodName}{CloseUpShot\xspace}

\newcommand{\Occlusionfilter}{Occlusion-aware Noise Suppression\xspace}
\newcommand{\occlusionfilter}{occlusion-aware noise suppression\xspace}
\newcommand{\globalguidance}{global structure guidance\xspace}
\newcommand{\GlobalGuidance}{Global Structure Guidance\xspace}

\usepackage[labelsep=period]{caption}
\renewcommand{\paragraph}[1]{\vspace{0.2em}\noindent \textbf{#1 \hspace{0.2em}}}

\begin{document}

\title{\MethodName: Close-up Novel View Synthesis from \\Sparse-views via Point-conditioned Diffusion Model}

\author{Yuqi Zhang, 
        Guanying Chen,~\IEEEmembership{Member,~IEEE,}
        Jiaxing Chen,
        Chuanyu Fu,\\
        Chuan Huang,~\IEEEmembership{Member,~IEEE,}
        Shuguang Cui,~\IEEEmembership{Fellow,~IEEE}

\thanks{Yuqi Zhang, Jiaxing Chen, Chuan Huang and Shuguang Cui are with the Shenzhen Future Network of Intelligence Institute (FNii-Shenzhen) and the Chinese University of Hong Kong at Shenzhen (CUHKSZ), China. Guanying Chen and Chuanyu Fu are with the Sun Yat-sen University, Shenzhen, China. Correspondence E-mail: chenguanying@mail.sysu.edu.cn}%
\thanks{Manuscript received April 19, 2021; revised August 16, 2021.}}

\markboth{Journal of \LaTeX\ Class Files,~Vol.~14, No.~8, August~2021}%
{Shell \MakeLowercase{\textit{et al.}}: A Sample Article Using IEEEtran.cls for IEEE Journals}

\IEEEpubid{0000--0000/00\$00.00~\copyright~2021 IEEE}

\twocolumn[{
\renewcommand\twocolumn[1][]{#1}
\maketitle

\begin{center}
    \vspace{-1em}
    \captionsetup{type=figure}
    \includegraphics[width=\textwidth]{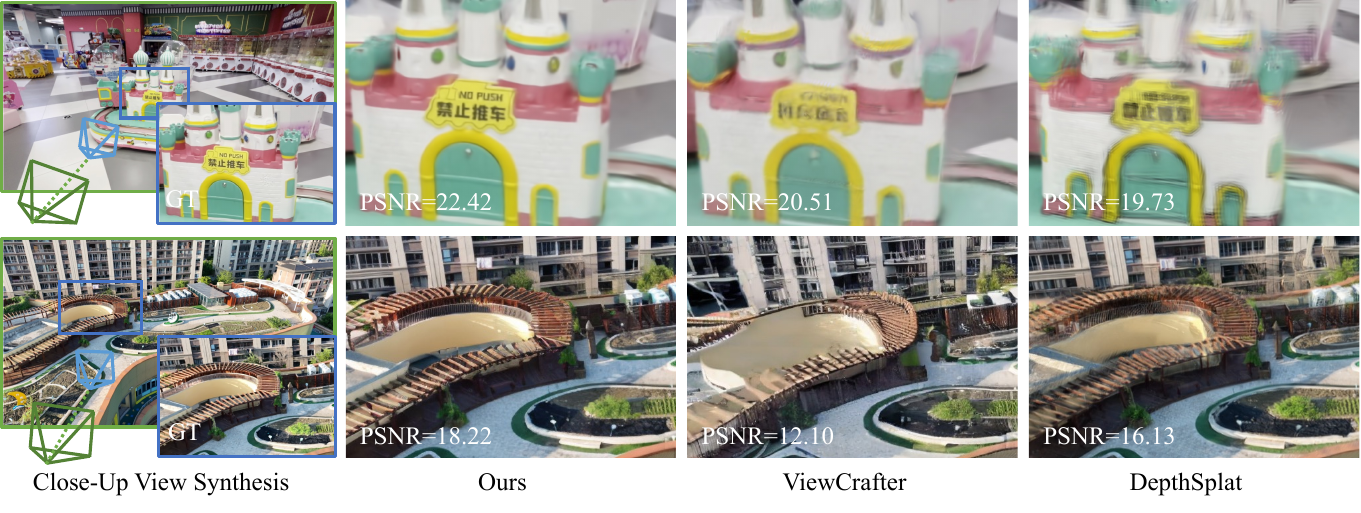}
    \\
    \vspace{-0.5em}
    \captionof{figure}{Given sparse-view inputs, we propose \MethodName, a novel-view synthesis framework that leverages diffusion prior to generate high-fidelity close-up images and support detail-preserving 3D reconstruction, especially when users move forward or zoom in (\eg, the original green camera move forward to the close-up blue camera in the left column) for fine-grained inspection.
    }
    \label{fig:teaser}
\end{center}
}]

\begin{abstract}
Reconstructing 3D scenes and synthesizing novel views from sparse input views is a highly challenging task.
Recent advances in video diffusion models have demonstrated strong temporal reasoning capabilities, making them a promising tool for enhancing reconstruction quality under sparse-view settings.
However, existing approaches are primarily designed for modest viewpoint variations, which struggle in capturing fine-grained details in close-up scenarios since input information is severely limited.
In this paper, we present a diffusion-based framework, called \MethodName, for close-up novel view synthesis from sparse inputs via point-conditioned video diffusion.
Specifically, we observe that pixel-warping conditioning suffers from severe sparsity and background leakage in close-up settings.
To address this, we propose \emph{hierarchical warping} and \emph{\occlusionfilter}, enhancing the quality and completeness of the conditioning images for the video diffusion model.
Furthermore, we introduce \emph{\globalguidance}, which leverages a dense fused point cloud to provide consistent geometric context to the diffusion process, to compensate for the lack of globally consistent 3D constraints in sparse conditioning inputs.
Extensive experiments on multiple datasets demonstrate that our method outperforms existing approaches, especially in close-up novel view synthesis, clearly validating the effectiveness of our design.
\end{abstract}

\href{https://zyqz97.github.io/CloseUpShot}{https://zyqz97.github.io/CloseUpShot}

\section{Introduction}
\label{sec:intro}

\IEEEpubidadjcol
\IEEEPARstart{S}{ignificant} advancements have been made in novel view synthesis and 3D reconstruction from multi-view images in recent years. 
In particular, neural rendering methods such as Neural Radiance Fields (NeRF)~\cite{mildenhall2020nerf} and 3D Gaussian Splatting (3DGS)~\cite{kerbl20233d} have led to major breakthroughs in photo-realistic view synthesis~\cite{chen2024pgsr, li2025mpgs, tang2025ivr}. 
Most of these methods rely on densely captured input views to ensure high-quality reconstruction and rendering. 
This reliance on dense input presents a significant limitation in real-world applications, where acquiring a large number of views may be impractical due to constraints in time, hardware, or accessibility.

\IEEEpubidadjcol
Beyond the dense-view setting, more practical scenarios involve either \emph{sparse-view} inputs or \emph{close-up} novel views, where fine scene structures must be synthesized from limited observations.
Early attemps~\cite{deng2022depthnerf, Niemeyer_2022_RegNeRF, Wang_2023_SparseNeRF} followed the optimization-based pipelines of NeRF or 3DGS, adapting them to sparse-view settings by incorporating additional supervision such as depth maps or multi-view feature correspondences. 
More recently, a shift has emerged toward data-driven, feed-forward approaches that aim to directly infer 3D representations from sparse images using models trained on a large amount of data. 
These methods often incorporate geometric priors, such as epipolar geometry~\cite{charatan2024pixelsplat}, cost volumes~\cite{chen2024mvsplat, xu2024depthsplat, fei2024pixel, tang2024hisplat, wang2024freesplat,yang2024depth}, or learned feature mappings~\cite{zhang2024gaussian, min2024epipolar, zhang2025transplat}, to aggregate multi-view information and enhance 3D understanding. 
By training across diverse scenes, they gain strong generalization capabilities and can reconstruct scenes efficiently without per-scene optimization.
Despite these advantages, sparse-view reconstruction remains fundamentally challenging. 
Due to the limited input information, such methods often struggle under wide baselines, severe occlusions, or novel view extrapolation.

With the recent success of diffusion models in generative vision tasks~\cite{zhang2024gbr, li2024nvcomposer, cai2024baking, ni2024recondreamer, huang2025part, zhu2024isolated}, researchers have begun exploring their use in 3D reconstruction~\cite{yu2024lm, paul2024gaussian,sargent2024zeronvs, chen2024liftimage3d, wu2024reconfusion,liu2024reconx,xing2024dynamicrafter}. 
In particular, video diffusion models have demonstrated strong temporal reasoning capabilities, which have been adapted for novel view synthesis under sparse input conditions~\cite{chen2024mvsplat360,liu20243dgs,yu2024viewcrafter,zhang2025high,wu2025difix3d}. 
To effectively guide the denoising process, video diffusion models rely on conditioning images that encode scene priors, which are processed into latent representations and concatenated with the noise latent.
For example, ViewCrafter~\cite{yu2024viewcrafter}, SplatDiff~\cite{zhang2025high}, and 3DGS-Enhancer~\cite{liu20243dgs} leverage 3D-aware conditioning inputs, such as point-projection images or 3DGS renderings.
This strategy has proven to be a promising direction for enhancing novel view synthesis in sparse-view scenarios.

\begin{figure}[t]
  \centering
    \includegraphics[width=\columnwidth]{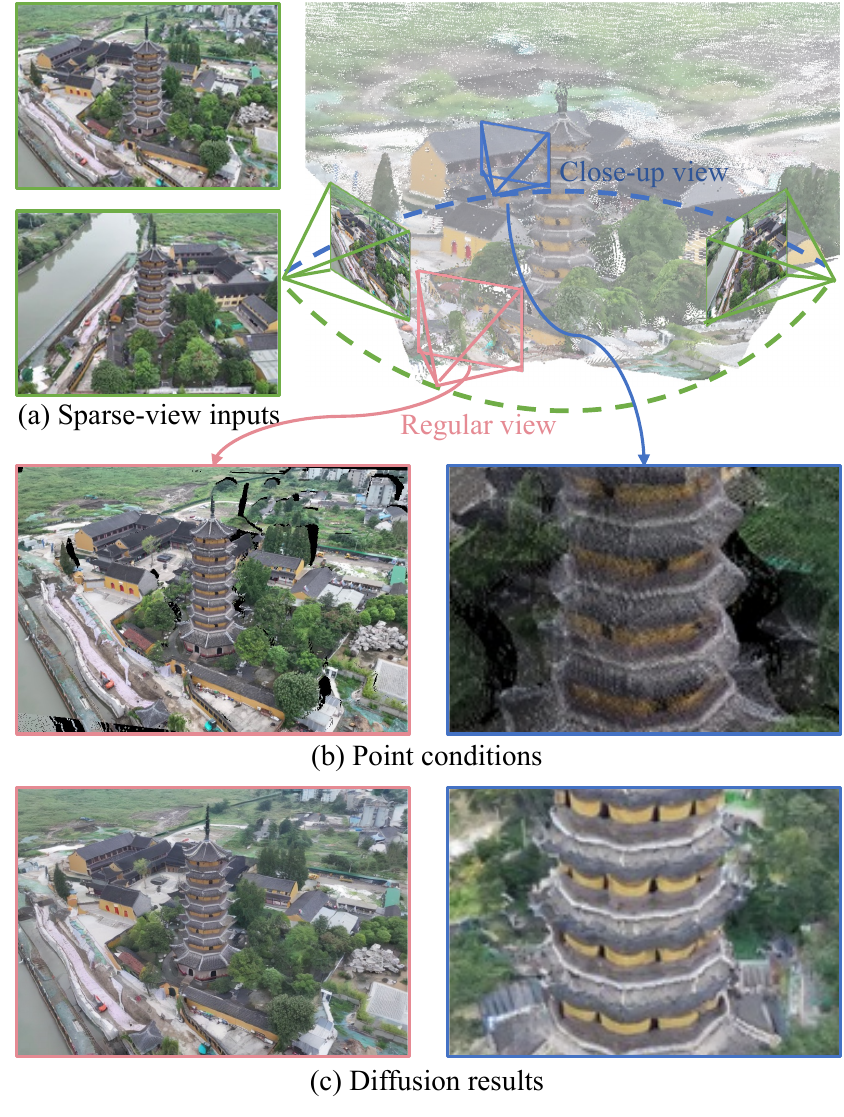}
    \vspace{-14pt}
    \\
    \caption{Limitations of point-conditioned diffusion models. (a) Given sparse input views, we extract point cloud, which is projected into a novel view to serve as conditioning (b) for the diffusion model (c). 
    When the target view is similar to the input views (e.g., the regular view), the projection is dense and offers effective guidance.
    However, for close-up views that require zooming in or moving closer, the projected conditioning becomes sparse and incomplete. These weak conditioning signals fail to guide the diffusion model effectively, leading to low-fidelity and artifact-prone outputs.}
  \label{fig:problem}
  \vspace{-14pt}
\end{figure}

However, most existing methods operate under the assumption of fixed camera intrinsics and modest view shifts.
In practical scenarios, such as free-viewpoint exploration, users often engage in \emph{close-up novel viewing} (see \fref{fig:teaser}), where they move forward or zoom in to inspect fine-grained scene details.
Such operations increase the sampling rate~\cite{yu2024mip} and exacerbate the inherent sparsity of sparse-view inputs, leading to more incomplete scene coverage. \zyqm{In the paper, we follow ViewCrafter~\cite{yu2024viewcrafter}, which adopts the point-conditioned diffusion model.
We find that the quality of the conditioning images plays a critical role in the performance of video diffusion models (see \fref{fig:problem}).}
Specifically, we observe that under close-up settings, sparse-view inputs inherently produce sparsely distributed 3D point clouds. 
When splatting into the novel view, these sparse point clouds result in incomplete conditioning images, leaving large holes and missing regions that fail to provide meaningful guidance for the diffusion process.
Furthermore, the sparsity of the 3D points also results in background leakage, where points from the background pass through gaps in the foreground. 
These leaked projections introduce error-prone conditioning signals, potentially misleading the generation process and resulting in visible artifacts. 
Moreover, depth maps predicted from sparse-view inputs inevitably suffer from view-dependent inconsistencies, \eg, a surface may receive conflicting depth values across different views. These inconsistencies result in noisy or contradictory conditioning inputs, degrading the generated results, particularly under close-up settings.

To tackle these challenges, we propose a framework, called \emph{\MethodName}, that improves sparse-view 3D reconstruction and novel view synthesis under the close-up setting. \zyqm{The key insight of our work is to enhance the quality of conditioning images, which in turn leads to better performance of the video diffusion model.}
Our method adopts a point-conditioned diffusion model and introduces three key design. 
First, a \emph{hierarchical warping} strategy that performs multi-resolution forward warping to produce dense conditioning images.
Second, we propose \emph{\occlusionfilter} that applies adaptive depth dilation to suppress background leakage. 
Third, we propose a \emph{\globalguidance} module that incorporates a consistent global 3D point cloud to provide unified geometric context for the diffusion model.

In summary, our key contributions are:
\begin{itemize}
    \item \zyqm{We propose a framework for close-up novel view synthesis and 3D reconstruction from sparse views. Our method achieves state-of-the-art performance on multiple datasets under the close-up setting, outperforming ViewCrafter by 28.7\% in PSNR on the DL3DV-10K dataset.}
    
    \item \zyqm{We introduce two effective modules, hierarchical warping for producing dender conditioning images, and occlusion-aware noise suppression for suppressing the leaked noise, that effectively address the limitations of point-splatting diffusion models under close-up settings.}
    
    \item We propose a global structure guidance mechanism that incorporates a unified and consistent 3D geometric context into the diffusion process, further improving the view consistency and structural fidelity.

\end{itemize}

\section{Related Work}
\label{sec:related_works}

\subsection{Feed-forward 3D Reconstruction}
Recently, a notable breakthrough, DUSt3R~\cite{wang2024dust3r}, introduced a Transformer-based architecture that directly learns from large-scale data to jointly predict point maps and camera poses in a single forward pass, drawing widespread attention. This work inspired a series of follow-up "3R" methods~\cite{wang20243d, leroy2024grounding, slam3r, zhang2024monst3r, lu2024align3r, wang2025continuous, vuong2025aerialmegadepth, yuan2025test3r}.

Several follow-up studies have extended the DUSt3R framework to handle large-scale image inputs, addressing its limitations in global optimization across multiple views~\cite{Yang_2025_Fast3R, cabon2025must3r}. 
Pow3R~\cite{jang2025pow3r} augmented DUSt3R with auxiliary information through a unified network architecture capable of processing multimodal inputs, thereby boosting performance. VGGT~\cite{wang2025vggt} introduced an alternating frame attention and global attention mechanism to jointly infer point maps, depth maps, and camera poses.

Forward 3D reconstruction methods based on 3DGS~\cite{kerbl20233d} have also emerged rapidly. 
On one hand, several approaches utilize camera poses predicted by DUSt3R to perform 3DGS-based reconstruction. Splatt3R~\cite{smart2024splatt3r} add a additional gaussian head opon MASt3R~\cite{leroy2024grounding} to predict the parameters of 3DGS. InstantSplat~\cite{fan2024instantsplat} first initializes point clouds and camera poses using MASt3R, then performs iterative optimization of both poses and 3DGS representations.
On the other hand, some methods aim to jointly optimize both camera poses and 3D Gaussian representations in a unified framework~\cite{chen2024pref3rposefreefeedforward3d,ye2024noposplat,kang2024selfsplat,li2024smilesplat, jiang2025anysplat}. FLARE~\cite{zhang2025flare} employs a two-stage strategy for both camera pose estimation and geometry reconstruction. This method first predicts 3DGS positions from two-stage optimization, while the remaining parameters are then estimated through a CNN network.

\subsection{Sparse-view 3D Reconstruction}
Under sparse-view inputs where Structure-from-Motion fails to recover accurate camera poses, both Neural Radiance Fields (NeRF)~\cite{mildenhall2020nerf} and 3DGS methods exhibit significant performance degradation. Most NeRF-based methods incorporate additional depth constraints to enhance reconstruction performance~\cite{deng2022depthnerf, Niemeyer_2022_RegNeRF, Wang_2023_SparseNeRF}. 
\zyqm{CoR-GS~\cite{zhang2024cor} proposes to train two separate 3DGS models and perform mutual regularization by comparing the disagreements between their output points and rendering, and CVT-xRF~\cite{zhong2024cvt} employs a voxel-based sampling strategy combined with a transformer to aggregate local regions, enhancing consistency among neighboring points. 
Moreover, You et al. ~\cite{you2023learning} propose a point-cloud-based framework that performs point cloud fusion before rendering, and further leverages the 3D geometry information to guide image restoration. 
}

Recently, several 3DGS-based methods for feed-forward sparse-view 3D reconstruction have explored diverse strategies. 
PixelSplat~\cite{charatan2024pixelsplat} leverages epipolar geometry priors to guide the splatting process. Other approaches construct cost volumes to aggregate cross-view information~\cite{chen2024mvsplat, xu2024depthsplat, fei2024pixel, tang2024hisplat, wang2024freesplat}. For instance, MVSplat~\cite{chen2024mvsplat} utilizes a transformer architecture to build a carefully designed cross-view cost volume, followed by a 2D U-Net that directly regresses Gaussian parameters. DepthSplat~\cite{xu2024depthsplat} further enhances this pipeline by incorporating a monocular depth prior~\cite{yang2024depth} to improve reconstruction quality. In parallel, some methods~\cite{zhang2024gaussian, min2024epipolar, zhang2025transplat} address the sparse-view challenge by exploiting image feature mapping to guide the reconstruction process.
However, these approaches depend on ground-truth camera poses, which are often unavailable in real-world sparse-view scenarios. 

\subsection{3D Reconstruction with Diffusion Prior} 
With the growing popularity of diffusion models, an increasing number of 3D reconstruction methods—particularly those focused on novel view synthesis—have begun to incorporate or fine-tune diffusion models to learn powerful priors~\cite{yu2024lm, zhang2024gbr, li2024nvcomposer, cai2024baking, ni2024recondreamer, paul2024gaussian,sargent2024zeronvs, chen2024liftimage3d, guo2025multi, xu2025geometrycrafter, wu2025video}. 
ReconFusion~\cite{wu2024reconfusion} first reconstructs the 3D scene using PixelNeRF~\cite{yu2021pixelnerf}, then extracts features from the rendered images as conditioning inputs to fine-tune a diffusion model. 
ReconX~\cite{liu2024reconx} leverages DUSt3R~\cite{wang2024dust3r} to extract a scene point cloud and introduces a 3D structure guidance mechanism to inject geometric information into a video diffusion model~\cite{xing2024dynamicrafter}. 
ViewExtrapolator~\cite{liu2024novel} proposes a training-free strategy inspired by the RePaint~\cite{lugmayr2022repaint} paradigm, performing guided and unguided denoising over different regions to eliminate out-of-distribution artifacts. 
\zyqm{NVS-Solver~\cite{you2025nvs} adaptively modulates the sampling process of pre-trained video diffusion model, enabling training-free novel view synthesis for both static and dynamic scenes.
}

To address the sparse-view input challenge, MVSplat360~\cite{chen2024mvsplat360} extends MVSplat~\cite{chen2024mvsplat} by directly rendering latent features using 3DGS as conditioning inputs for Stable Video Diffusion (SVD). 
\zyqm{Zhong et al.~\cite{zhong2025taming} proposes a training-free strategy that directly controls diffusion model to generate consistent images using 3DGS-rendered images as guidance, and has been shown to be effective in indoor scenes.}
3DGS-Enhancer~\cite{liu20243dgs} simulates low-quality GS renderings under sparse-view settings, which are then used to adapt the SVD model via fine-tuning, effectively mitigating artifacts caused by sparse observations. 
\zyqm{
Instead of conditioning directly on 3DGS-rendered images, several approaches~\cite{yu2024viewcrafter,zhang2025high,ren2025gen3c,ma2025you,cao2025mvgenmaster} adopt warp-based conditioning strategies.}
ViewCrafter~\cite{yu2024viewcrafter} proposes a point-based representation approach, utilizing point-projection maps as conditions to fine-tune the SVD model~\cite{xing2024dynamicrafter}. 
\zyqm{GEN3C~\cite{ren2025gen3c} employs a spatial-temporal 3D cache that fuses multi-view features via max-pooling, incorporating visibility awareness to handle view-dependent effects.
See3D~\cite{ma2025you} also adopts warp-image conditioning to eliminate the reliance on explicit pose control in video diffusion, achieving effective 3D reconstruction and generation across various types of scenes. 
MVGenMaster~\cite{cao2025mvgenmaster} adpots 3D priors, including the warped RGB images, as condition and can generate up to 100 novel views with one forward process.
}
Furthermore, Difix3D+~\cite{wu2025difix3d} leverages a diffusion prior during both training and inference for enhanced novel view synthesis, and employs a single-step SD-Turbo~\cite{sauer2024adversarial} backbone to improve computational efficiency.

Nevertheless, the aforementioned methods on sparse-view setting are predominantly designed for regular view distributions. 
In contrast, close-up novel views further amplify the inherent challenges of sparse-view settings. 
Such scenarios remain largely underexplored, despite being crucial for applications requiring fine-grained 3D understanding.

\begin{figure*}[t] \centering
    \label{fig:overview}
    \includegraphics[width=\textwidth]{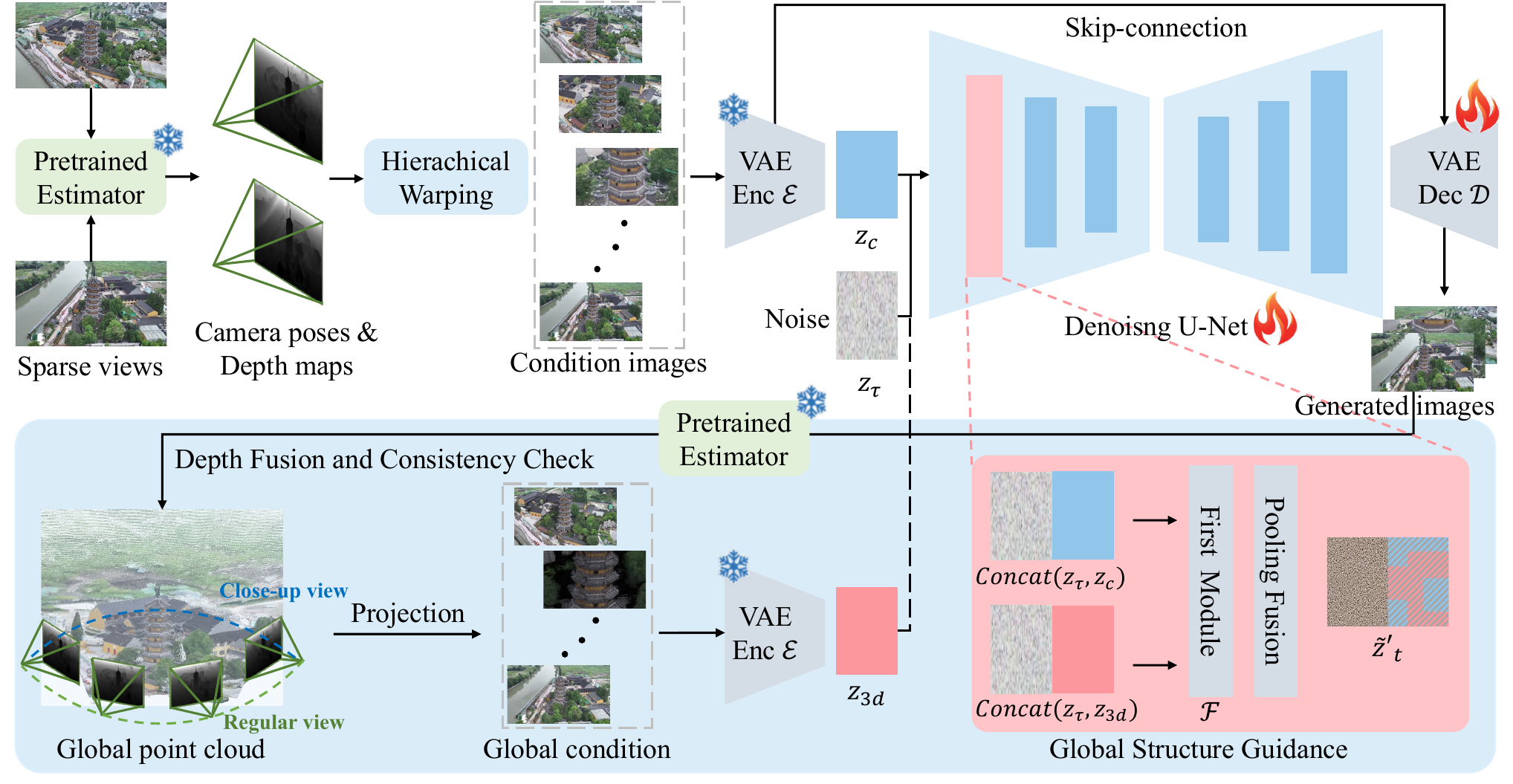}
    \caption{Overview. Our pipeline takes two sparse input views and is capable of synthesizing fine-grained novel views under close-up settings using a point-conditioned video diffusion model.
    First, a pretrained estimator is applied to obtain depth maps and camera parameters from the input images.
    Second, we introduce two effective modules, hierarchical warping and \occlusionfilter, to enhance the sparse and noisy conditioning images, especially in the close-up setting.
    Third, we perform a multi-view consistency check to construct a global point cloud, which is projected into target views to provide global structure guidance for the denoising U-Net.
    Finally, the generated novel views, together with the reference inputs, are used to supervise 3DGS for photorealistic and detail-preserving 3D reconstruction.
    } 
    \label{fig:pipeline}
\end{figure*}

\section{Method}
\label{sec:method}

Given extremely sparse input views, this work tackles the problem of high-quality 3D reconstruction by leveraging diffusion priors to improve close-up novel view synthesis. 
Our approach builds upon recent advances in 3D Gaussian Splatting (3DGS) optimization, where projected point clouds are used as conditioning inputs to guide a video diffusion model~\cite{xing2024dynamicrafter}. 
This projection-based conditioning provides dense guidance when the sampling rate~\cite{yu2024mip} of the novel view close to the reference views. 
However, its effectiveness diminishes when the sampling rate changes in close-up setting, \eg, increasing the focal length or decreasing the distance between camera and scene. In these cases, the projected points become sparse and unreliable, leading to poor detail synthesis. 
On the other hand, the feed-forward 3DGS methods reconstruction without diffusion prior fail to recover fine details due to insufficient information from sparse-view inputs.

To address these limitations, we introduce three key improvements (see \fref{fig:pipeline}). First, a hierarchical warping strategy enhances the spatial coverage of conditioning images across multiple scales (in \sref{sec:hierarchical_warping}) that guide the video diffusion process. 
Second, we develops an \occlusionfilter mechanism to address visibility-related artifacts caused by sparse projections, \eg, background point leakage through foreground gaps (in \sref{sec:noise_suppress}).
Third, we apply a \globalguidance mechanism that provides unified and consistent global geometric context to the diffusion model to enhance the consistency of 3D reconstruction
(in \sref{sec:global}). 
Together, these components enable our method to reconstruct high-fidelity 3D Gaussian Splatting representations ((in \sref{sec:3dgs})) from as few as two input views, significantly improving the visual quality of close-up novel view renderings.

\subsection{Preliminary: Video Diffusion Model}
\label{sec:SVD_model}

A diffusion model consists of a forward process $q$ and a reverse process $p_\theta$~\cite{song2021ddim}. 
The forward process $q(\mathbf{x}_t \mid \mathbf{x}_{t-1}, t)$ gradually corrupts clean latent representations $\mathbf{x}_0$ by adding Gaussian noise over time steps $t = 1, \ldots, T$. 
The reverse process approximates the denoising trajectory using a neural network $\epsilon_\theta$, which removes noise from $\mathbf{x}_t$ to recover the original latent. A video diffusion framework consists of three components: a VAE encoder $\mathcal{E}$, a VAE decoder $\mathcal{D}$, and a U-Net-based denoising network $\epsilon_\theta$. 
During training, the ground-truth video $\mathbf{x} \in \mathbb{R}^{L \times 3 \times H \times W}$, where $L$ is the number of frames, is first encoded into a latent representation $\mathbf{z} = \mathcal{E}(\mathbf{x}) \in \mathbb{R}^{L \times C \times h \times w}$. The diffusion process is then applied in this latent space.

In this paper, we adopt an Image-to-Video (I2V) Diffusion Model~\cite{xing2024dynamicrafter}. 
To guide the generation, we employ two types of conditioning: 1) CLIP~\cite{radford2021clip} features extracted from the input reference image are used in the U-Net via cross-attention; 2) a condition latents $\mathbf{z}_c$, which has the same spatial-temporal dimensions as $\mathbf{z}$, is concatenated with the noisy latent $\mathbf{z}_{\tau}$ along the channel dimension as the denoising input:

\begin{equation}
  \label{eq:latent_concat}
  \tilde{\mathbf{z}_t} = \text{Concat}(\mathbf{z}_{\tau}, \mathbf{z}_c).
\end{equation}
The denoising network $\epsilon_\theta$ predicts the noise added at each timestep based on $\tilde{\mathbf{z}_t}$, the timestep $t$, and the CLIP condition. The training objective minimizes the mean squared error between the predicted noise and the ground true noise:
\begin{equation}
\mathcal{L}_{\theta} = \mathbb{E}_{t, \mathbf{z}_0, \boldsymbol{\epsilon}} \left[ \left\| \boldsymbol{\epsilon} - \epsilon_\theta(\tilde{\mathbf{z}_t}, t, \text{CLIP}(\cdot)) \right\|_2^2 \right].
\end{equation}

At inference time, we adopt the DDIM~\cite{song2021ddim} sampling strategy with classifier-free guidance~\cite{ho2021classifier} to iteratively denoise random Gaussian noise to recover a clean latent representation $\hat{\mathbf{z}}_0$.%
The final video is then obtained by the VAE decoder: $\hat{\mathbf{x}} = D(\hat{\mathbf{z}}_0)$.

\subsection{Hierarchical Warping for Diffusion Conditioning}
\label{sec:hierarchical_warping}
Our work bulids opon ViewCrafter~\cite{yu2024viewcrafter}, which trains a video diffusion model conditioned on a combination of reference images and point-projection renderings. 
Given sparse-view inputs, ViewCrafter first employs DUSt3R~\cite{wang2024dust3r} to estimate depth maps and camera parameters. 
These depth maps are then back-projected into 3D space to form point clouds, which are reprojected into the target view using PyTorch3D, yielding point-based conditioning images. These conditioning frames, together with the sparse input images, are fed into a video diffusion model to guide the denoising process.
ViewCrafter demonstrates effectiveness when the target view maintains a similar sampling rate~\cite{yu2024mip}, \eg, the focal length and camera distance remain similar to reference views, yielding dense point-projection renderings for reliable guidance. 

However, under more challenging scenarios when capturing more details of scenes (close-up setting), such as zooming in or reducing the camera-to-object distance, the sparsity of the input point cloud becomes a limiting factor. The projected conditioning images tend to contain large empty regions, offering unreliable guidance to the denoising model.
Moreover, feed-forward 3DGS reconstruction methods without diffusion prior, such as DepthSplat~\cite{xu2024depthsplat}, inherently struggle in infering the fine details due to the limited content from sparse observed inputs. 
As a result, they tend to exhibit blurry or visible artifacts, particularly in fine-detail regions.

\noindent\textbf{Hierarchical warping.} 
To overcome these challenges, we leverage the generative power of diffusion models to hallucinate plausible scene details and propose a hierarchical warping strategy that performs pixel-level forward warping at multiple spatial resolutions. 
Given two sparse-view images as input (also refer to reference images $I_{r}$), we first estimate the per-frame depth $D_r$ and camera pose $P_r$ using VGGT~\cite{wang2025vggt}, a recent efficient feed-forward pointmap estimation framework.

\begin{figure}[t]
  \centering
    \includegraphics[width=\columnwidth]{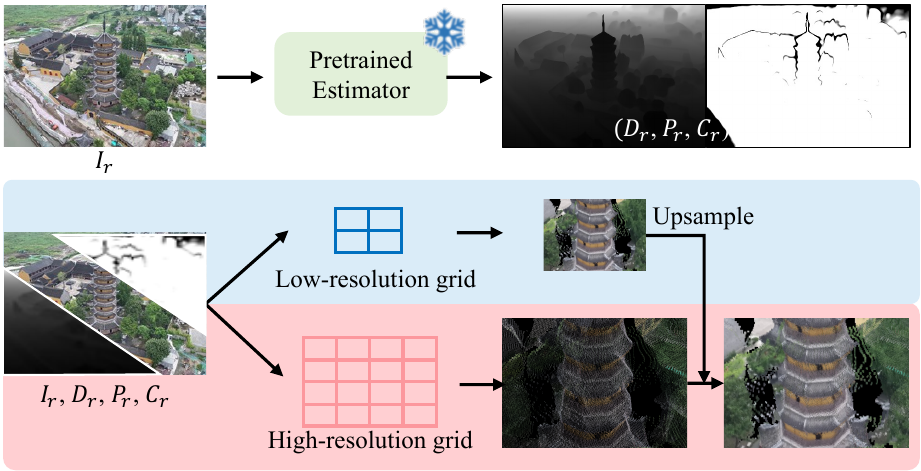}
    \caption{Hierarchical Warping for Diffusion Conditioning. We perform forward warping at both high and low resolutions to obtain a sharp but sparse high-resolution image and a blurry but dense low-resolution image. The low-resolution result is then upsampled to fill the missing regions in the high-resolution image, producing a dense conditioning input for the diffusion model. Note that we only illustrate the reliable regions for simplicity.}
  \label{fig:hier_warping}
  \vspace{-12pt}
\end{figure}

The core insight of our method lies in a hierarchical warping strategy that operates at multiple resolutions to produce more complete and occlusion-respecting conditioning images. 
We perform forward warping~\cite{jin2025flovd} from both input views into the target view $P_t$ using the predicted depth and camera parameters. 
At the original resolution, a simple forward warping leads to a sparse conditioning image $I_t^{\text{high}}$, especially when capturing fine scene details, as previously discussed.
To alleviate this, we downsample the target grid and perform warping from high-resolution reference images to the low-resolution target grid $I_t^{\text{low}}$. 
This produces a dense low-resolution warping result that inherently accounts for occlusions, and provides more reliable RGB values, as illustrated in \fref{fig:hier_warping}.
Crucially, this approach maintains visibility gaps caused by occlusions rather than naively interpolating over them. 
The dense result is then upsampled and used to fill missing regions in the original-resolution warping image:
\begin{equation}
I_t^{\text{high} / \text{low}} = \text{Warp}(I_r, D_r, P_r \rightarrow P_t^{\text{high} / \text{low}}),
\end{equation}

\begin{equation}
I_t = \mathbbm{1}(I_t^{\text{high}}) \cdot I_t^{\text{high}} + (1 - \mathbbm{1}(I_t^{\text{high}})) \cdot \text{Upsample}(I_t^{\text{low}}),
\end{equation}
where $\mathbbm{1}(\cdot)$ denotes the indicator of the valid pixel. Moreover, the warping results from different reference images may meet conflicts, we merge these by retaining pixels from the reference view that is closer to the target camera center. 

\noindent\textbf{Confidence-aware reliability division.} Furthermore, we observe that the depth predictions from VGGT may contain errors, which can be amplified by our hierarchical warping strategy. 
To mitigate this issue, we introduce a confidence-guided partitioning scheme for warping. 
Specifically, we adopt the depth prediction head of VGGT to estimate both a depth map and a corresponding confidence map for each input reference view. 
Based on the confidence values, we divide the depth map into reliable and unreliable regions: the top 90\% of pixels (by confidence) are regarded as reliable, while the bottom 10\% are labeled as unreliable.
In addition, we find that VGGT tends to produce inaccurate depth estimates around object boundaries. 
To account for this, we apply gradient-based edge detection and treat boundary regions as unreliable as well.

\begin{algorithm}[t]
\caption{Hierarchical Warping for Diffusion Conditioning}
\label{alg:hierarchical_warping}
\textbf{Input: } Reference images $\{I_r\}, r = 0, 1, ...,R$ ($R$ denotes the number of reference images)\\
\textbf{Output: } Hierarchically warped conditioning images $\{I_t\}$ under target poses $\{P_t\}$
\begin{algorithmic}[1]
\STATE Use pretrained estimator to predict depths $D_r$, confidence maps $C_r$, and camera poses $P_r$ from $I_r$: $\{D_r, C_r, P_r\} = \text{Estimator}(\{I_r\})$
\FOR{each target view $P_t$}
\FOR{each reference view $r \in R$}
    \STATE Divide $D_r$ into reliable $D_r^{\text{reliable}}$ and unreliable regions $D_r^{\text{unrel}}$ based on $C_r$ and edge gradients
    \STATE \textbf{// High-resolution warping on reliable regions}
    \STATE $I_t^{\text{high}} \leftarrow \text{Warp}(I_r, D_r^{\text{reliable}}, P_r \rightarrow P_t^{\text{high}})$
    \STATE \textbf{// Low-resolution warping on reliable regions}
    \STATE Downsample target grid $I_t^{\text{low}}$, and corresponding poses $P_t^{\text{low}}$
    \STATE $I_t^{\text{low}} \leftarrow \text{Warp}(I_r, D_r^{\text{reliable}}, P_r \rightarrow P_t^{\text{low}})$
    \STATE \textbf{// Combine the results of two resolution}
    \STATE $I_t^{\text{reliable}} = \mathbbm{1}(I_t^{\text{high}}) \cdot I_t^{\text{high}} + (1 - \mathbbm{1}(I_t^{\text{high}})) \cdot \text{Upsample}(I_t^{\text{low}})$

    \STATE \textbf{// Repeat for unreliable regions}
    \STATE $I_t^{\text{high-ur}} \leftarrow \text{Warp}(I_r, D_r^{\text{unrel}}, P_r \rightarrow P_t^{\text{high}})$
    \STATE $I_t^{\text{low-ur}} \leftarrow \text{Warp}(I_r, D_r^{\text{unrel}}, P_r \rightarrow P_t^{\text{low}})$
    \STATE \resizebox{\linewidth}{!}{$I_t^{\text{unrel}} = \mathbbm{1}({I_t^{\text{high-ur}}}) \cdot I_t^{\text{high-ur}} + (1 - \mathbbm{1}(I_t^{\text{high-ur}})) \cdot \text{Upsample}(I_t^{\text{low-ur}})$}
    \STATE \textbf{// Merge results of the two regions}
    \STATE ${I_t}^{r} = \mathbbm{1}(I_t^{\text{reliable}}) \cdot I_t^{\text{reliable}} 
+ (1 - \mathbbm{1}(I_t^{\text{reliable}})) \cdot I_t^{\text{unrel}}$
\ENDFOR
\STATE  $I_t = \text{Merge}(\{{I_t}^r\}), r = 0, 1, ..., R$
\ENDFOR
\STATE \textbf{return} $\{I_t\}$
\end{algorithmic}
\end{algorithm}

With this reliability devision, our hierarchical warping strategy is executed in a two-stage process. First, we perform high-resolution forward warping using only the reliable depth points, followed by a low-resolution warping stage that fills in the missing regions. 
This coarse warping stage leverages denser spatial coverage at lower resolution to produce plausible RGB values while preserving occlusion-induced holes. 
Next, the same two-stage warping procedure is applied to the unreliable regions to complete the remaining image.
By prioritizing reliable depth regions, our hierarchical warping strategy mitigates the adverse effects of depth uncertainty and provides cleaner, more informative conditioning inputs for the diffusion model. Please refer to Algorithm~\ref{alg:hierarchical_warping} for the whole algorithm.

\subsection{\Occlusionfilter}
\label{sec:noise_suppress}

\begin{figure}[t]
  \centering
  \includegraphics[width=0.48\columnwidth]{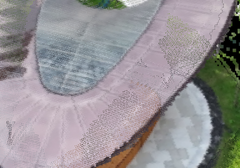}
  \includegraphics[width=0.48\columnwidth]{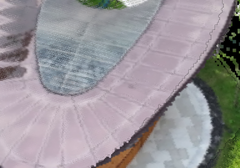}
  \\
  \includegraphics[width=0.48\columnwidth]{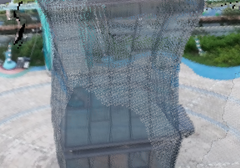}
  \includegraphics[width=0.48\columnwidth]{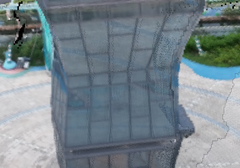}
  \\
  \vspace{-1pt}
  \makebox[0.48\columnwidth]{\footnotesize Background leakage}
  \makebox[0.48\columnwidth]{\footnotesize Noise suppression}
  \\
  \caption{Problem of background leakage. When close-up viewing, background points often leak through gaps in the sparse foreground, leading to incorrect projections conditioning images. Our noise suppression strategy mitigates this issue by filtering out these artifacts, resulting in more reliable and cleaner conditioning images for diffusion generation.}
  \label{fig:filter}
\end{figure}

When capturing fine-grained scene details in close-up setting, another challenge arises: occlusion artifacts caused by sparse point projections. 
On the one hand, due to point cloud sparsity, pixel warping-based diffusion models may produce artifacts where background content ``leaks through'' the gaps between foreground points, as shown in \fref{fig:filter}. 
On the other hand, 3DGS-based methods also struggle in such settings.  
For instance, when increasing the focal length or moving the camera closer to the scene, densification often fails to provide sufficient Gaussian primitives in these fine-detail regions, leading to occlusion violations as well.
A straight-forward solution might be to render multi-view images at the same sampling rates, which can mitigate occlusion artifacts. 
However, these synthesized views often remain blurry or noisy because of the insufficient scene information provided by the limited input views.

To tackle this, we propose an \occlusionfilter strategy. 
We observed that background points tend to leak through more severely when the warped projection is more sparse. 
Thus, we adopt a dynamic dilation strategy that adaptively expands depth map based on the sparsity of the warped depth map.

Concretely, for each target view, we first obtain the depth map $D_{\text{warp}}$ directly using the warping strategy at high-resolution described in \sref{sec:hierarchical_warping}. 
Based on the density, which denotes the proportion of valid pixels in $D_{\text{warp}}$, we then dynamically determine the dilation kernel size: the sparser the warping, the larger the dilation kernel. 
This yields a dilated depth map $D_{\text{dilate}}$ that preserves original depths under dense warping, while applying depth expansion under sparse warping to suppress background leakage.
Next, we filter out background projections that violate the depth consistency with:
\begin{equation}
M_{\text{occ}}(\mathbf{p}) = 
\begin{cases}
0, & \text{if } D_{\text{warp}}(\mathbf{p}) - D_{\text{dilate}}(\mathbf{p}) > \tau_D \\
1, & \text{otherwise}
\end{cases}
\end{equation}
where $\tau_D = 0.2$ is a loose threshold that preserves true foreground-background separation while tolerating depth estimation noise. \zyqm{Please kindly refer to the supplementary material for more details and qualitative analysis.}

Importantly, we apply this noise suppression only to regions with high depth confidence, since low-confidence depths are more error-prone and may degrade performance.
Moreover, we apply this noise suppression only during inference, since overly clean training conditions diminish the input diversity required to stimulate robust learning.

\subsection{\GlobalGuidance}
\label{sec:global}
While our proposed \textit{hierarchical warping} and \textit{\occlusionfilter} strategies effectively enhance the conditioning images and thereby improve the model’s ability to handle close-up novel views, they remain inherently local and rely on per-view warping results.
In practice, the depth maps from pretrained estimators often exhibit inevitable inconsistencies across different reference views, \eg, the same geometric structure may be assigned different depth values depending on the viewpoint, leading to inconsistent conditioning images after forward warping.
These inconsistencies in the conditioning inputs can propagate into the denoising process, resulting in view-inconsistent generations and degraded performance in 3D reconstruction and novel view synthesis. 
To address this issue, we introduce a \textit{\globalguidance} strategy that provides a unified geometric context for the diffusion model.

\noindent\textbf{Global point cloud generation via DepthFusion.} 
First, we leverage our previously described strategies to finetune a video diffusion model, obtaining a coarse model capable of synthesizing novel views. 
The images generated by the coarse model are then fed into VGGT to predict per-frame camera parameters and depth maps. 
With these attributes, we perform a global consistency check based on DepthFusion~\cite{cheng2020deep}, producing a unified global point cloud.
Specifically, for a given view, we adopt the warping strategy from DepthFusion to find pixel-wise correspondences across multiple source views. For each pixel on the given view, we obtain its 3D point $\mathbf{p}_i$ and a set of corresponding points ${\mathbf{p}_j}$ from other views. We determine geometric consistency by measuring the Euclidean distance. However, we find that relying solely on geometric checks is insufficient and produces clutter points. Thus, we introduce an additional color consistency constraint to improve the global point cloud quality. The global consistency check can be described as below:

\begin{equation}
  \label{eq:df_check}
    M(\mathbf{p}_i, \mathbf{p}_j, \mathbf{c}_i, \mathbf{c}_j) = \|\mathbf{p}_i - \mathbf{p}_j\|_2 < \tau_{\text{g}} \quad \& \quad \|\mathbf{c}_i - \mathbf{c}_j\|_2 < \tau_{\text{c}} ,
\end{equation}
where $i, j \in {0, 1, \ldots, V}$, with $V$ denoting the number of frames in the diffusion model. Here, $\mathbf{p}$ and $\mathbf{c}$ represent the 3D position and color value. Then, if a point has more than $\tau_{\text{num}}$ consistent correspondences, we refine its 3D position by averaging the matched points. The threshold $\tau_{\text{g}}, \tau_{\text{c}}, \tau_{\text{num}}$ is set to 0.01, 0.1, 10, respectively.

\noindent\textbf{Structure guidance denoising.}
After obtaining the global point cloud, we project it into each reference view to generate a multi-view consistent conditioning image $I_{\text{g}}$.
We inject this global geometric context into the diffusion model through a pooling fusion strategy. While GEN3C~\cite{ren2025gen3c} also employs a pooling strategy to fuse multi-view point-projection images and incorporates mask visibility into the model, our method differs by introducing a globally consistent 3D point cloud as geometric guidance.
As described in \sref{sec:SVD_model}, the conditioning image obtained from our hierarchical warping is first encoded by a VAE encoder into a latent $\mathbf{z}_c$, which is then concatenated with the noise latent $\mathbf{z}_\tau$ to form $\tilde{\mathbf{z}_t}$, the input to the denoising U-Net.
Similarly, the projected consistent image $I_{\text{g}}$ is processed by the same VAE encoder to obtain its latent $\mathbf{z}_{\text{g}}=\mathcal{E}(I_{\text{g}})$. which is also concatenated with $\mathbf{z}_\tau$ to form $\tilde{\mathbf{z}_t}^{\text{g}}$.
Both the concatenated latents are passed through the first module $\mathcal{F}$ of the U-Net for feature upsampling. We then apply max pooling over the two feature maps to fuse the local and global geometric signals, which is fed into the subsequent modules of the U-Net. The fusion process can be described as:

\begin{equation}
  \label{eq:pooling_fusion}
    \tilde{\mathbf{z}_t}' = \mathrm{Maxpool}\Big(\mathcal{F}(\tilde{\mathbf{z}_t}), \mathcal{F}(\tilde{\mathbf{z}_t}^{\text{g}})\Big).
\end{equation}

\noindent\textbf{Decoder finetune.}
Moreover, the features extracted by the VAE encoder contain rich spatial details and texture information, which can be beneficial for improving generation fidelity.
To leverage this, we introduce skip connections from the VAE encoder to the decoder. 
During this stage, we freeze both the U-Net backbone and the VAE encoder, and fine-tune only the decoder layers.
This strategy allows the model to better reconstruct fine-grained textures, leading to improved visual fidelity in the generated results.

\begin{figure}[tbp]
  \centering
    \includegraphics[width=\columnwidth]{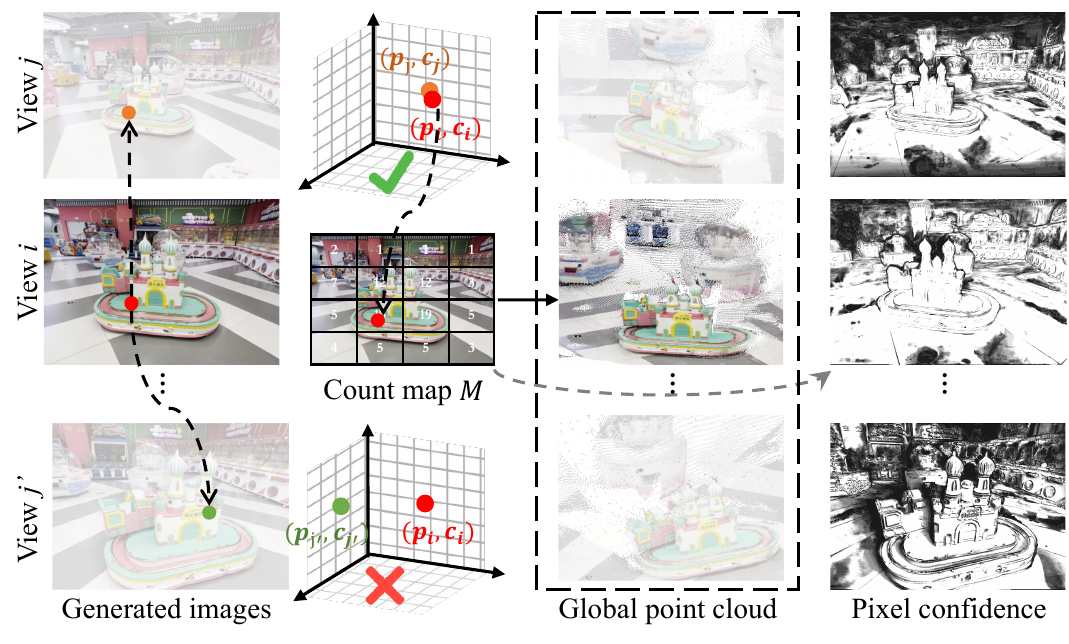}
    \caption{Global point cloud generation and pixel-level confidence map. 
    We perform geometric and photometric consistency checks between each generated view and all other views. 
    For each pixel, we count the number of views which passes the consistency check. A global point cloud is then obtained by thresholding these counts and aggregating consistent pixels across views. In addition, the pixel-level confidence map can be computed from the count map directly.}
  \label{fig:global_check}
\end{figure}

\subsection{3DGS Reconstruction}
\label{sec:3dgs}
\noindent\textbf{Vallina 3DGS optimization.}
After fine-tuning the video diffusion model, we obtain a set of generated images $I_{\text{gen}}$ from sparse input views. 
These generated views improve scene coverage and recover fine-grained details that are typically missing under sparse-view settings, especially when targeting close-up views.
To reconstruct the 3D geometry, we adopt 3D Gaussian Splatting (3DGS)~\cite{kerbl20233d}, which represents the scene as a collection of Gaussian primitives. 
Each primitive is parameterized by a learnable mean position $\boldsymbol{\mu}_i$, covariance matrix $\boldsymbol{\Sigma}_i$ (defining shape and orientation), opacity $\alpha_i$, and spherical harmonics coefficients $\mathbf{c}_i$ that model view-dependent color appearance.
3DGS enables fast rasterization by projecting these primitives onto the 2D image plane, allowing for photorealistic and real-time rendering.
The generated images are used as supervision for the 3DGS-rendered views, and the optimization minimizes a combination of L1 and SSIM losses:
\begin{equation}
\mathcal{L}_{\text{3dgs}} = (1 - \lambda) \cdot \mathcal{L}_1 + \lambda \cdot \mathcal{L}_{\text{SSIM}}.
\end{equation}

However, our experiment reveals that directly applying the vanilla 3DGS optimization with generated views often yields suboptimal results.
This is primarily due to the inherent inconsistencies stem from the video diffusion model, which is optimized for temporal synthesis rather than strict view-consistency.
These inconsistencies can degrade the quality of the reconstructed 3D geometry when optimizing 3DGS model.

\noindent\textbf{Confidence-aware optimization.}
To address this, we adopt a confidence-aware optimization strategy inspired by 3DGS-Enhancer~\cite{liu20243dgs}, which incorporates both image-level and pixel-level confidence maps to modulate the contribution of each supervision signal.
For the image-level confidence, we assign weights $W_\text{image}$ based on the geometric distance between each generated view and the input reference views, with closer views receiving higher confidence.

For pixel-level confidence, our method diverges from 3DGS-Enhancer. Instead of relying on projected scaling parameters of 3D Gaussians to estimate visibility, we leverage our previously proposed consistency checking strategy.
Specifically, for each generated view, we maintain a per-pixel count map $M$ that records how many other views are consistent with this pixel through geometric and photometric consistency checks in \eref{eq:df_check}.
The pixel-wise confidence is then computed as:
\begin{equation}
W_{\text{pixel}} = \min \left( \frac{M}{\tau_{\text{num}}},\ 1 \right),
\end{equation}
and then the loss function can be described as:

\begin{equation}
\mathcal{L}_{\text{3dgs}} = W_\text{image} \cdot \Big( W_\text{pixel}\cdot (1 - \lambda) \cdot \mathcal{L}_1 + \lambda \cdot \mathcal{L}_{\text{SSIM}}\Big).
\end{equation}

\begin{figure*}[t]
  \centering
    \includegraphics[width=\textwidth]{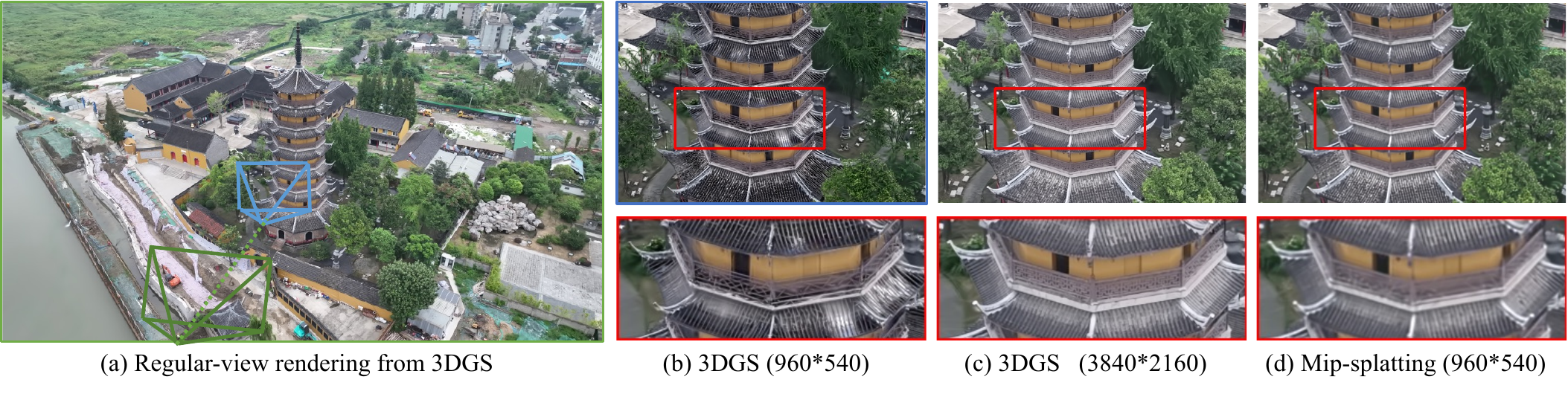}
    \caption{Close-up view ground-truth generation. (a) We first train a 3DGS model using medium-resolution images ($960 \times 540$), which produces satisfactory renderings under regular-view settings. However, when rendering close-up novel views, the model exhibits degraded results, as shown in (b). To address this, we train the model using $4\times$ resolution images, resulting in higher-fidelity ground truth renderings in (c). Red boxes highlight the improvements in rendering details, \eg, the railing on the tower. Additionally, we consider adopting Mip-splatting in (d), a method designed for handling view variations, but it still produces blurry results under close-up settings. Hence, we take the rendering from (c) as ground truth for training and evaluation.}
  \label{fig:discussion}
  \vspace{-6pt}
\end{figure*}

\section{Experiments}
\label{sec:Experiments}

\subsection{Training Dataset Construction.}
\noindent\textbf{Datasets.} 
We train and evaluate our method on the DL3DV-10K~\cite{ling2024dl3dv} and DL3DV-Drone~\cite{ling2024dl3dv} dataset, which provides 10,510 casually captured real-world scenes and 105 drone-captured scenes, respectively. 
For training, we employ a mixed-data strategy, using 1,000 scenes from the '3K' subset of DL3DV-10K dataset and the 100 scenes from DL3DV-Drone dataset.
We partition each scene into multiple segments and select the first and last frames of each segment as reference frames. After rigorous data filtering, we construct a dataset comprising approximately 2,000 video clips, each containing 25 frames.

\noindent\textbf{Close-up View Ground-truth Generation.} 
One major challenge is that neither DL3DV-10K nor DL3DV-Drone dataset provides ground truth for close-up novel views. To enable quantitative evaluation, we generate pseudo ground-truth for close-up view synthesis using high-quality 3D reconstructions.

\Fref{fig:discussion} presents an example illustrating the effectiveness of our pseudo ground truth generation process. We first train a 3DGS model using medium-resolution images ($960\times540$). 
As shown in (a), the rendering under the regular view appears satisfactory. 
However, when rendering novel close-up views, the results degrade significantly, as illustrated in (b). This degradation can be attributed to the lack of high-frequency details at medium-resolution, where the Gaussian primitives tend to be needle-shaped and only fit the regular viewpoints, resulting in visible artifacts when close-up view.
(c) To address this issue, we train 3DGS model using $4\times$ higher resolution images, which produces sharper and more reliable close-up renderings. We also adopt Mip-splatting~\cite{yu2024mip} in (d), a method designed for view variation scenarios, which use medium-resolution images for training that is more sufficient, but still produces blurry results. Hence, we take the rendeings from high-resolution 3DGS model as pseudo ground truth for evaluation, which provides reliable and effective images for assessing close-up novel view synthesis.

Moreover, while DL3DV-10K dataset provides both multi-view images and COLMAP reconstruction files, DL3DV-Drone contains only raw aerial videos without camera parameters. To prepare DL3DV-Drone for training and evaluation, we extract video frames to form a 360-degree image set and apply COLMAP to estimate camera poses.

\begin{table*}[!t] \centering
    \captionof{table}{Quantitative evaluation of sparse-view novel view synthesis on the DL3DV-10K dataset, including the close-up view and regular view. The asterisk * denotes that we enhance the close-up view synthesis using the reconstructed 3DGS from the regular view. The best result is highlighted in \textbf{bold}, and the second-best is \underline{underlined}, \zyqm{excluding Ours (DUSt3R) for better comparison.}}
    \label{table:dl3dv10k}
    
\resizebox{\textwidth}{!}{
\begin{tabular}{lcccccc|cccccc}
\toprule[0.15em] 
& \multicolumn{6}{c|}{{Close-up View}} & \multicolumn{6}{c}{{Regular View}}   \\ 
\cmidrule(lr){2-7} \cmidrule(lr){8-13}
& \multicolumn{3}{c}{Easy Set} & \multicolumn{3}{c|}{Hard Set}
& \multicolumn{3}{c}{Easy Set} & \multicolumn{3}{c}{Hard Set}
\\
\multirow{-3}{*}{Method} 
& PSNR $\uparrow$ & SSIM $\uparrow$ & LPIPS $\downarrow$
& PSNR $\uparrow$ & SSIM $\uparrow$ & LPIPS $\downarrow$
& PSNR $\uparrow$ & SSIM $\uparrow$ & LPIPS $\downarrow$
& PSNR $\uparrow$ & SSIM $\uparrow$ & LPIPS $\downarrow$
\\ \midrule
MVSplat~\cite{chen2024mvsplat}  
& 14.97 	& 0.480 	& 0.563 	
& 14.73 	& 0.498 	& 0.600 	
& 15.32 	& 0.442 	& 0.534 	
& 14.07 	& 0.429 	& 0.596 
\\
MVSplat360~\cite{chen2024mvsplat360}  
& 14.23 	& 0.456 	& 0.560 	
& 13.72 	& 0.469 	& 0.588 	
& 15.10 	& 0.446 	& 0.528 	
& 14.48 	& 0.449 	& 0.568 
\\
DepthSplat~\cite{xu2024depthsplat}  
& 17.93 	& 0.582 	& 0.454 	
& 16.44 	& 0.543 	& 0.522 	
& 19.46 	& 0.613 	& 0.390 	
& 15.92 	& 0.503 	& 0.501 
\\
DepthSplat*
& 18.65 	& \underline{0.607} 	& 0.474 	
& 16.68 	& 0.563 	& 0.549  
& - & - & -  
& - & - & -
\\
ViewCrafter~\cite{yu2024viewcrafter}  
& 13.82 	& 0.427 	& 0.533 	
& 13.06 	& 0.419 	& 0.567 	
& 18.65 	& 0.559 	& 0.342 	
& 16.40 	& 0.487 	& 0.438 
\\
ViewCrafter*
& 18.44 	& 0.589 	& 0.450 	
& 17.02 	& \underline{0.570} 	& 0.518 
& - & - & -  
& - & - & -
\\

\color{black}GEN3C~\cite{ren2025gen3c}
& \color{black}15.79 	& \color{black}0.481 	& \color{black}0.472 	
& \color{black}14.66 	& \color{black}0.463 	& \color{black}0.544 	
& \color{black}20.75 	& \color{black}\underline{0.678} 	& \color{black}\textbf{0.280}
& \color{black}17.51 	& \color{black}0.552 	& \color{black}\underline{0.400} 
\\
\color{black}Difix3D+~\cite{wu2025difix3d}
& \color{black}\underline{18.73} 	& \color{black}0.597 	& \color{black}\underline{0.356} 	
& \color{black}\underline{17.12} 	& \color{black}0.537 	& \color{black}\underline{0.412}
& \color{black}\textbf{21.15} 	& \color{black}0.665 	    & \color{black}\underline{0.288}
& \color{black}\textbf{18.83} 	& \color{black}\underline{0.564} 	& \color{black}\textbf{0.378}
\\
\midrule
Ours (VGGT)
& \textbf{20.61} 	& \textbf{0.688} 	& \textbf{0.342} 	
& \textbf{18.96} 	& \textbf{0.652} 	& \textbf{0.406} 	
& \underline{20.93} 	& \textbf{0.707} 	& {0.339} 	
& \underline{18.41} 	& \textbf{0.619} 	& {0.433} 
\\ 
\color{black}Ours (DUSt3R)
& \color{black}20.44 	& \color{black}0.669  	& \color{black}0.357 	
& \color{black}18.67 	& \color{black}0.618 	& \color{black}0.430  
& \color{black}20.72 	& \color{black}0.684     & \color{black}0.351 
& \color{black}18.28 	& \color{black}0.603     & \color{black}0.452 
\\

\bottomrule[0.15em]
\end{tabular}
}

    \centering
    \vspace{6pt}
    \makebox[0.160\textwidth]{\footnotesize Point Cloud Render}
    \makebox[0.160\textwidth]{\footnotesize DepthSplat}
    \makebox[0.160\textwidth]{\footnotesize ViewCrafter *}
    \makebox[0.160\textwidth]{\footnotesize Difix3D+}
    \makebox[0.160\textwidth]{\footnotesize Ours}
    \makebox[0.160\textwidth]{\footnotesize Ground Truth}
    \\
    \includegraphics[width=0.160\textwidth,height=0.102\textwidth]{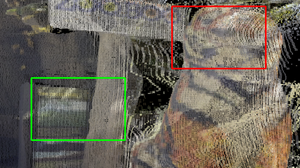}
    \includegraphics[width=0.160\textwidth,height=0.102\textwidth]{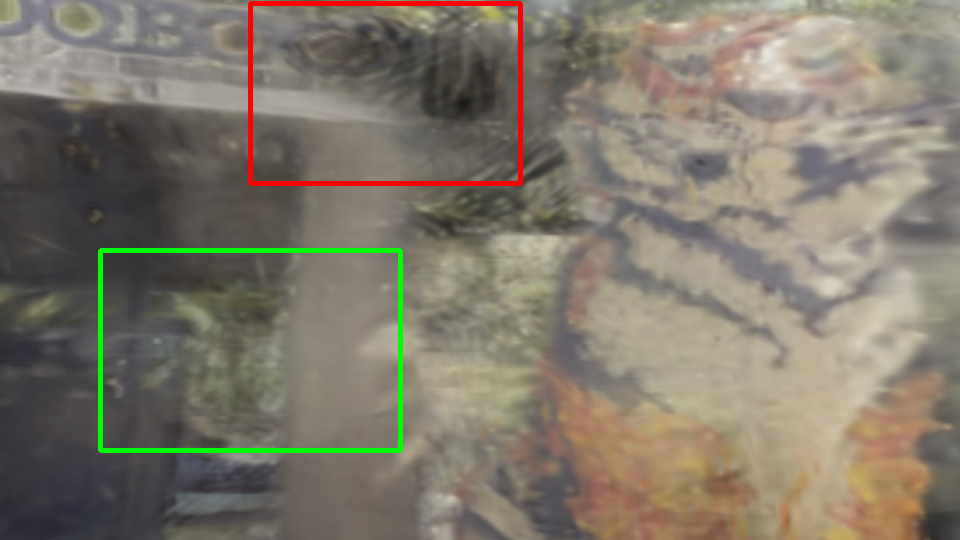}
    \includegraphics[width=0.160\textwidth,height=0.102\textwidth]{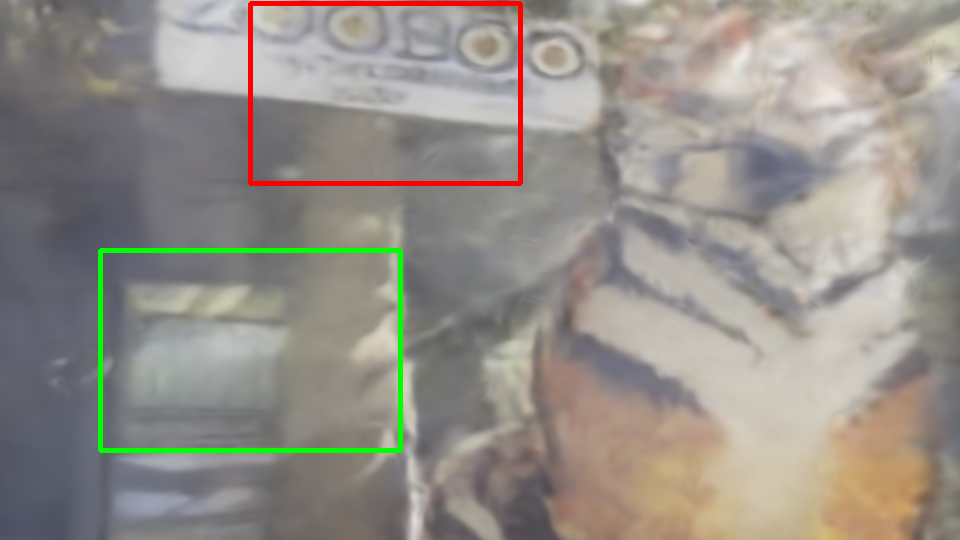}
    \includegraphics[width=0.160\textwidth,height=0.102\textwidth]{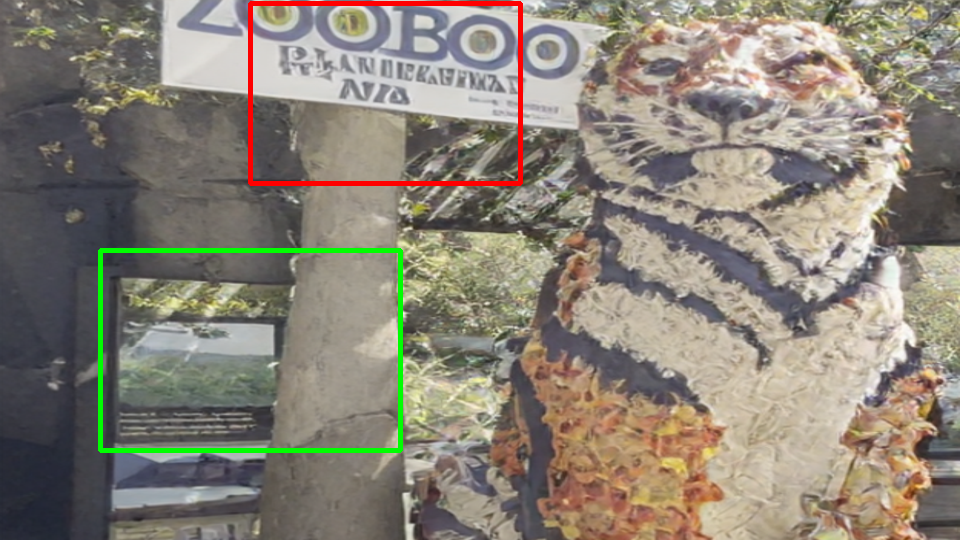}
    \includegraphics[width=0.160\textwidth,height=0.102\textwidth]{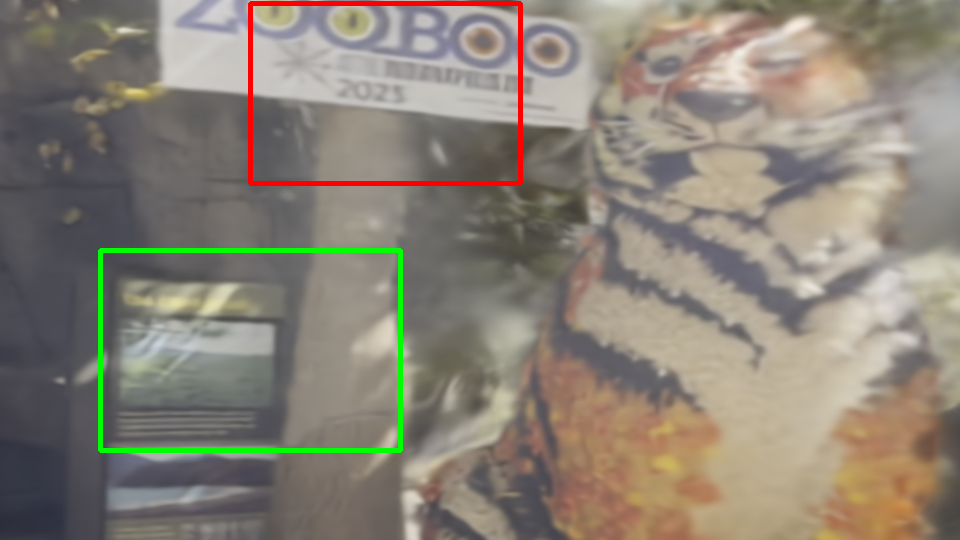}
    \includegraphics[width=0.160\textwidth,height=0.102\textwidth]{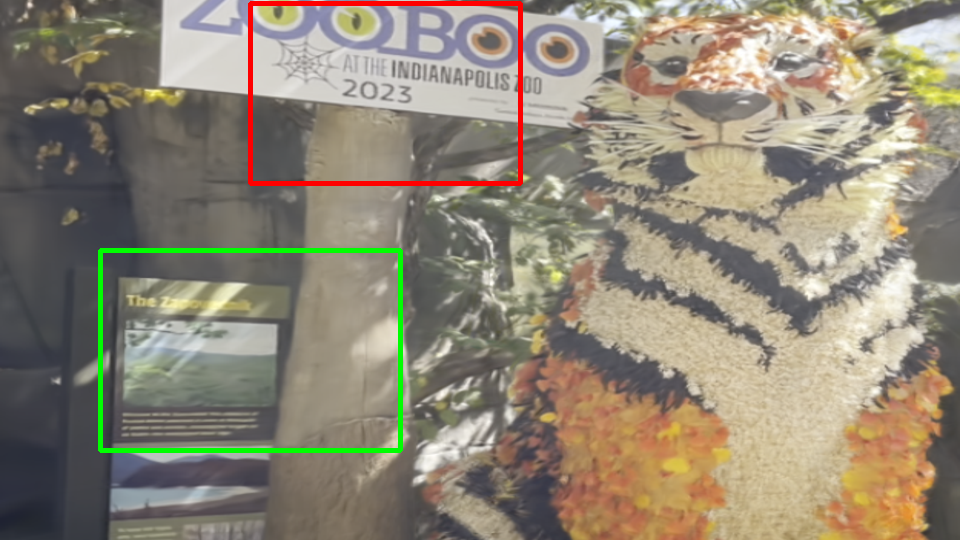}
    \\
    \includegraphics[width=0.0773\textwidth,height=0.056\textwidth]{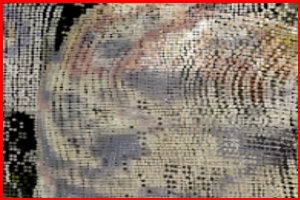}
    \includegraphics[width=0.0773\textwidth,height=0.056\textwidth]{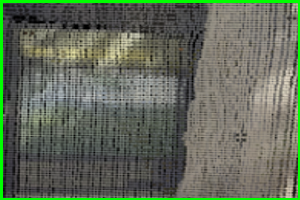}
    \includegraphics[width=0.0773\textwidth,height=0.056\textwidth]{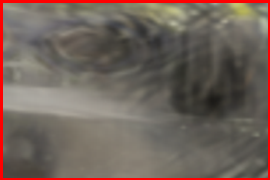}
    \includegraphics[width=0.0773\textwidth,height=0.056\textwidth]{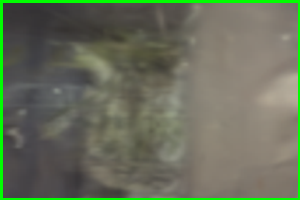}
    \includegraphics[width=0.0773\textwidth,height=0.056\textwidth]{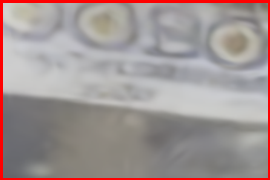}
    \includegraphics[width=0.0773\textwidth,height=0.056\textwidth]{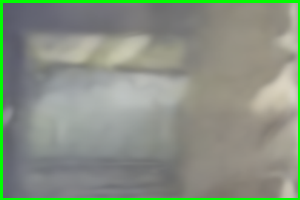}
    \includegraphics[width=0.0773\textwidth,height=0.056\textwidth]{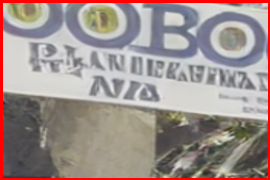}
    \includegraphics[width=0.0773\textwidth,height=0.056\textwidth]{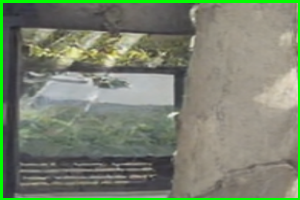}
    \includegraphics[width=0.0773\textwidth,height=0.056\textwidth]{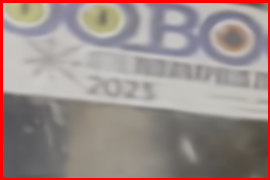}
    \includegraphics[width=0.0773\textwidth,height=0.056\textwidth]{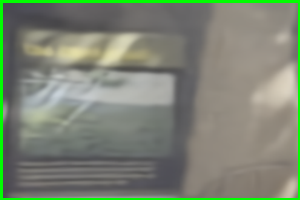}
    \includegraphics[width=0.0773\textwidth,height=0.056\textwidth]{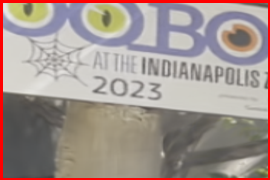}
    \includegraphics[width=0.0773\textwidth,height=0.056\textwidth]{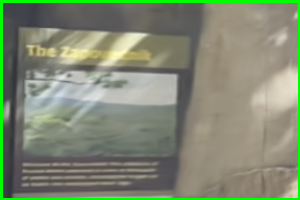}
    \\
    \vspace{6pt}
    \includegraphics[width=0.160\textwidth,height=0.102\textwidth]{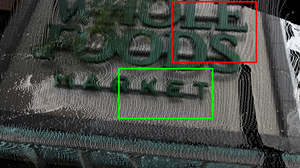}
    \includegraphics[width=0.160\textwidth,height=0.102\textwidth]{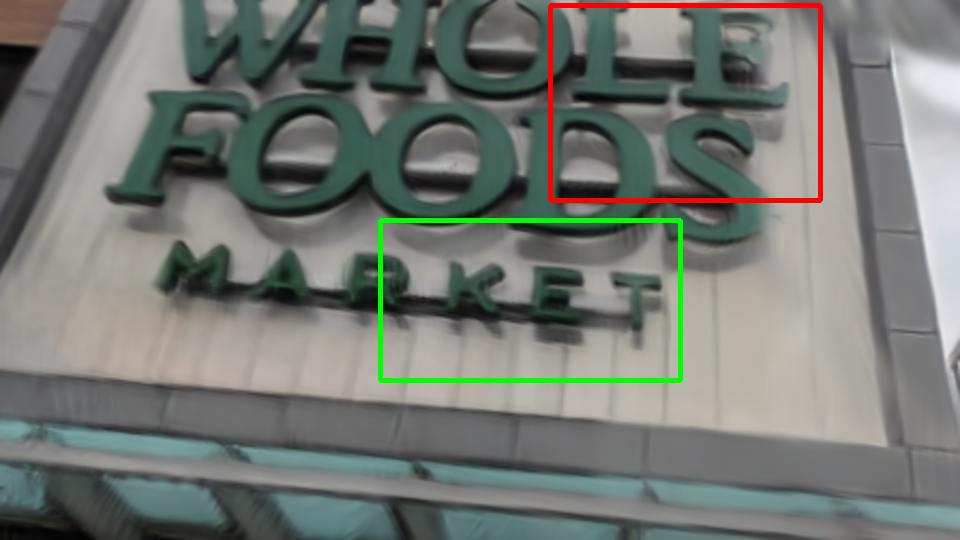}
    \includegraphics[width=0.160\textwidth,height=0.102\textwidth]{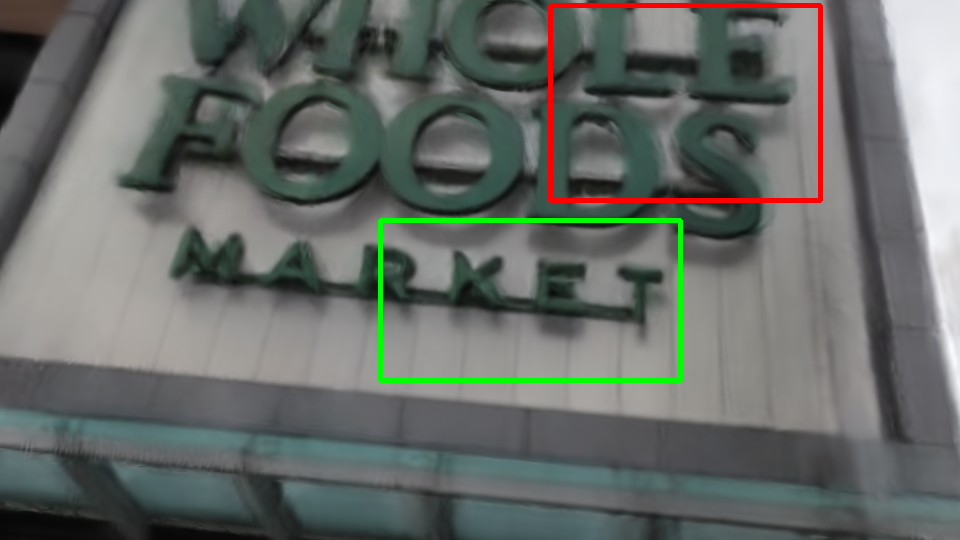}
    \includegraphics[width=0.160\textwidth,height=0.102\textwidth]{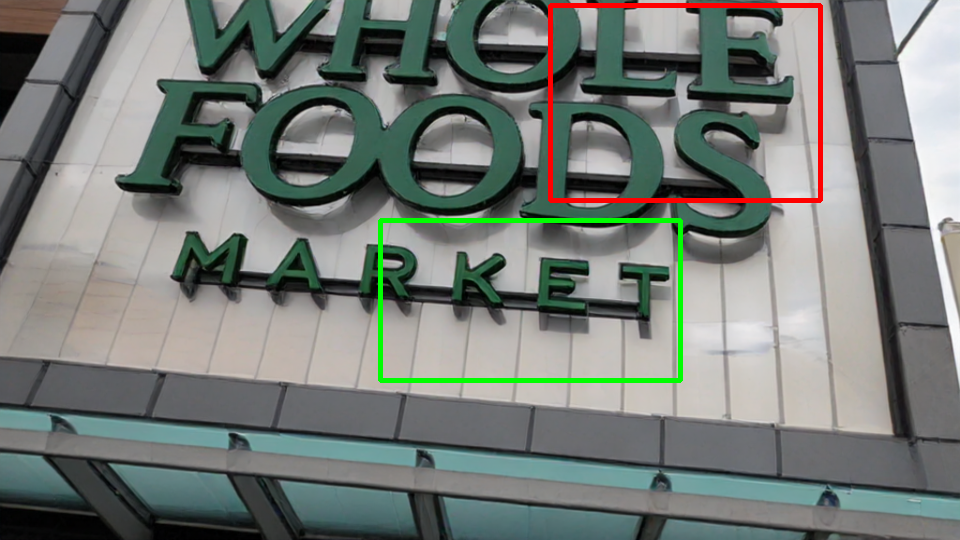}
    \includegraphics[width=0.160\textwidth,height=0.102\textwidth]{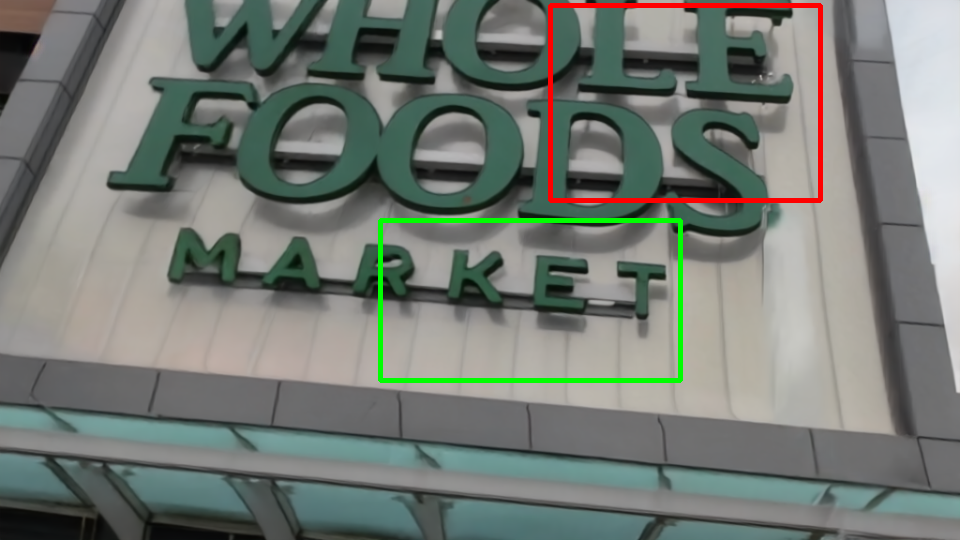}
    \includegraphics[width=0.160\textwidth,height=0.102\textwidth]{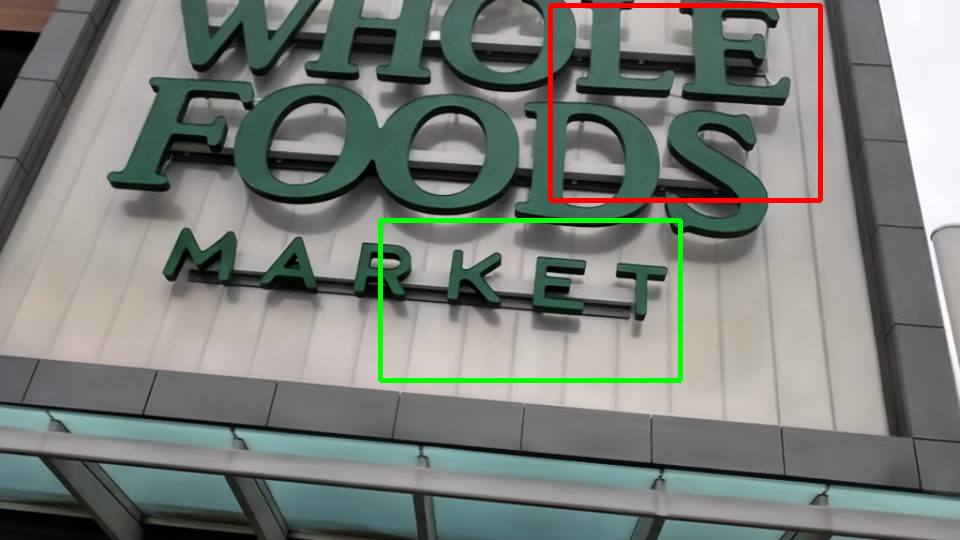}
    \\
    \includegraphics[width=0.0773\textwidth,height=0.056\textwidth]{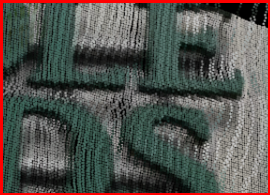}    
    \includegraphics[width=0.0773\textwidth,height=0.056\textwidth]{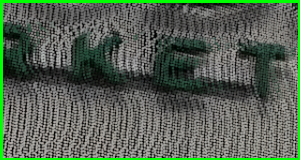}
    \includegraphics[width=0.0773\textwidth,height=0.056\textwidth]{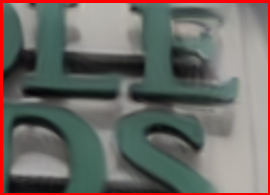}
    \includegraphics[width=0.0773\textwidth,height=0.056\textwidth]{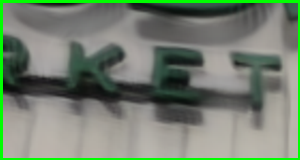}
    \includegraphics[width=0.0773\textwidth,height=0.056\textwidth]{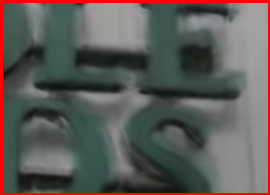}
    \includegraphics[width=0.0773\textwidth,height=0.056\textwidth]{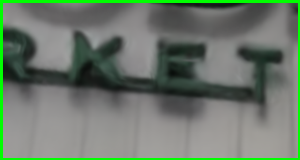}
    \includegraphics[width=0.0773\textwidth,height=0.056\textwidth]{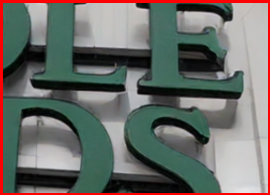}
    \includegraphics[width=0.0773\textwidth,height=0.056\textwidth]{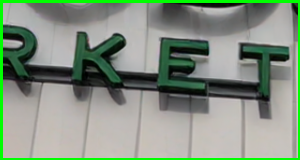}
    \includegraphics[width=0.0773\textwidth,height=0.056\textwidth]{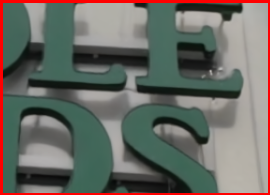}
    \includegraphics[width=0.0773\textwidth,height=0.056\textwidth]{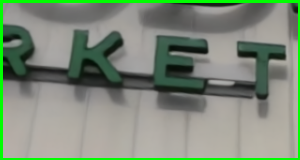}
    \includegraphics[width=0.0773\textwidth,height=0.056\textwidth]{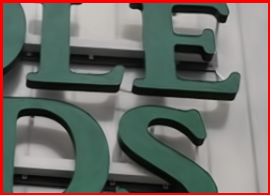}
    \includegraphics[width=0.0773\textwidth,height=0.056\textwidth]{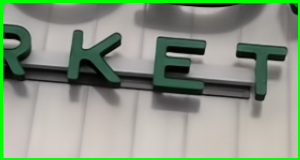}

    \captionof{figure}{Qualitative results of close-up novel view synthesis on the DL3DV-10K dataset. The color boxes highlight the difference among the methods for better comparison.}
    \label{fig:dl3dv10k_a}
\end{table*}

\subsection{Implementation Details}
We use the video diffusion model proposed by DynamiCrafter~\cite{xing2024dynamicrafter}, and fine-tune it based on the sparse-view weights provided by ViewCrafter~\cite{yu2024viewcrafter}. 
We train our video diffusion pipeline at a resolution of 1024×576 with a 25-frame video sequence.
First, we train the video denoising U-Net using the proposed hierarchical warping images for 5000 iterations with a batch size of 4.
Next, we freeze the U-Net weights and fine-tune the decoder by introducing skip connections from the encoder features, also for 5000 iterations with a batch size of 4.
Once this training stage is complete, we generate multi-view images from the U-Net to construct a global point cloud, obtaining the global structure context.
We then apply the proposed global structure guidance and re-train the video denoising U-Net from scratch for 5000 iterations.
As before, we freeze the U-Net and further fine-tune the decoder in the final stage. 
The complete training process requires 3 days in total, with the UNet backbone trained for 24 hours at a fixed learning rate of $1\times10^{-5}$ and the decoder fine-tuning taking 12 hours at a learning rate of $1\times10^{-4}$.
We apply the DDIM~\cite{song2021ddim} sampler with classifier-free guidance~\cite{ho2022classifier} during inference. Finally, the generated multi-view images are used to supervise 3D Gaussian Splatting, which is trained for 1000 iterations per scene.

\noindent\textbf{Baseline Methods.} 
We compare our approach with several representative and the state-of-the-art baselines designed for novel view synthesis under sparse inputs. These include 3DGS-based methods such as MVSplat~\cite{chen2024mvsplat} and DepthSplat~\cite{xu2024depthsplat}, as well as diffusion-based methods like MVSplat360~\cite{chen2024mvsplat360}, \zyqm{GEN3C}~\cite{ren2025gen3c}, and ViewCrafter~\cite{yu2024viewcrafter}. 
MVSplat, MVSplat360, and DepthSplat use ground-truth camera poses during training and inference, while our method and ViewCrafter rely on pretrained estimators to predict camera parameters.
For all baselines, we adopt the official implementations. The DepthSplat model we use is trained in a two-stage manner, which is first pre-trained on the RealEstate10K~\cite{zhou2018stereo}dataset and then fine-tuned on the DL3DV datasets. 
MVSplat is trained solely on RealEstate10K, while MVSplat360 builds upon MVSplat by incorporating a Stable Video Diffusion module and is further fine-tuned on DL3DV dataset. 
ViewCrafter is trained on a mixture of DL3DV and RealEstate10K datasets. 
\zyqm{For GEN3C, we adopt the released implementation namely 'the video generation from multi-view images' based on NVIDIA Cosmos, and use VGGT as pretrained estimator to obtain the cameras parameteres and depth maps for fair comparison. Difix3D+~\cite{wu2025difix3d} is trained on the randomly selected 112 scenes from total 140 scenes on the DL3DV-Benchmark. We follow the official implementation of Difix3D+, which adoptes progressive 3D updating by Difix3D and further uses post render processing by Difix3D+.}
All experiments are conducted under the two-view input setting for fair comparison.

\noindent\textbf{Evaluation.} 
For evaluation, we randomly select 8 scenes from DL3DV-Benchmark (a subset of DL3DV-10K) and 5 scenes from DL3DV-Drone. All these evaluation scenes were removed from our training dataset to prevent any overlap. \zyqm{For close-up view simulation, we enlarge the camera focal length by a factor of 4$~\sim~$5 or move the cameras closer to the scene by 50$~\sim~$60\% of the maximum image depth. 
Moreover, we classify the test scenes in DL3DV-Benchmark into an easy set and a hard set, which depends on the spacing between reference images, to assess the performance under varying conditions. Please refer to supplementary material for more details.}

\noindent\textbf{Metrics.} 
To measure the performance of the proposed method, we use the pixel-aligned metrics: Peak Signal-to-Noise Ratio (PSNR) and Structural Similarity Index Measure (SSIM)~\cite{wang2004image}, as well as the perceptual metric: Learned Perceptual Image Patch Similarity (LPIPS)~\cite{johnson2016perceptual}.

\begin{figure*}[tbp]
  \centering
    \input{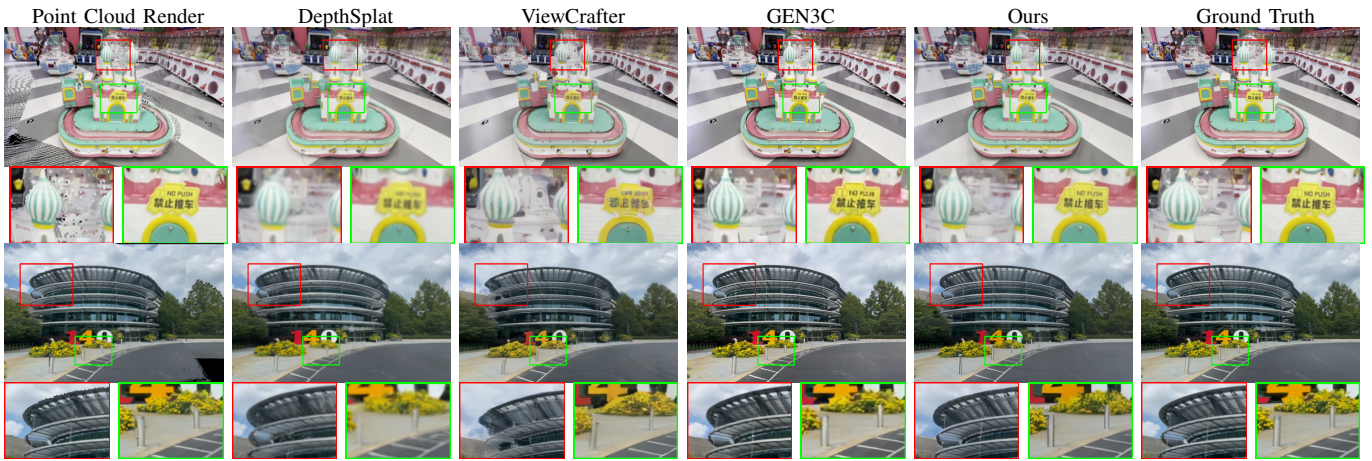}
    \caption{Qualitative results of regular-view novel view synthesis on the DL3DV-10K dataset. The color boxes highlight the difference among the methods for better comparison.}
  \label{fig:dl3dv10k_b}
\end{figure*}

\subsection{Evaluation on DL3DV-10K Datasets}
We first evaluate our method on the DL3DV-10K dataset. 
\Tref{table:dl3dv10k} presents quantitative results for both close-up and regular views synthesis.
Considering that baseline methods generally perform poorly on close-up views, we additionally provide enhanced versions of DepthSplat and ViewCrafter for comparison, which are denoted with an asterisk (*). 
Specifically, we train a 3DGS model using their regular-view outputs and use this model to render close-up views.

\zyqm{For the regular views, our method significantly outperforms ViewCrafter, and achieves a competitive performance compared with the state-of-the-art method, GEN3C~\cite{ren2025gen3c} and Difix3D+~\cite{wu2025difix3d}, obtaining the best and second-best scores in PSNR and SSIM.
On the more challenging close-up view, our method demonstrates clear advantages with outperforming metrics.
Specifically, Our method achieves a PSNR of 20.61 db and a SSIM of 0.688 on the easy set, outperforming the second-best method by approximately 2 dB and 0.08, respectively. 
More notably, it yields a lower LPIPS score (0.342), indicating superior perceptual quality. 
The improvements on the hard set are also significant, which highlights its robustness under challenging viewing conditions.
\fref{fig:dl3dv10k_a} and \fref{fig:dl3dv10k_b} illustrate the qualitative comparison under close-up views and regular views, respectively. Our method achieves more plausible results in the close-up view setting, \eg,our method accurately reconsructed the text on the signboard, while Difix3D+ generates overall sharper but incorrect textures. 
In the regular view, our approach also preserves finer details, including clearer text regions and sharper building structures.}

\begin{table}[t]
  \caption{Quantitative evaluation of sparse-view novel view synthesis on the DL3DV-Drone dataset. The asterisk * denotes that we enhance the close-up view synthesis using the reconstructed 3DGS from the regular view.
  }
  \label{table:drone}
  \resizebox{0.49\textwidth}{!}{
\begin{tabular}{lccc|ccc}
\toprule[0.15em]
\multirow{2}{*}{Method} & \multicolumn{3}{c|}{Close-up View} & \multicolumn{3}{c}{Regular View} \\
\cmidrule(lr){2-4} \cmidrule(lr){5-7}
& PSNR$\uparrow$ & SSIM$\uparrow$ & LPIPS$\downarrow$ 
& PSNR$\uparrow$ & SSIM$\uparrow$ & LPIPS$\downarrow$ \\
\midrule
MVSplat~\cite{chen2024mvsplat}  
& 14.87 	& 0.411 	& 0.628 	
& 14.70 	& 0.315 	& 0.612 
\\
MVSplat360~\cite{chen2024mvsplat360}  
& 14.32 	& 0.401 	& 0.622 	
& 15.09 	& 0.334 	& 0.614 
\\

DepthSplat~\cite{xu2024depthsplat}  
& 17.59 	& 0.489 	& 0.455 	
& 18.40 	& 0.471 	& 0.376 
\\
ViewCrafter~\cite{yu2024viewcrafter} 
& 15.41 	& 0.369 	& 0.524 	& 18.76 	& 0.423 	& 0.358 
\\
ViewCrafter*
& 17.95 & 0.472 & 0.519 & - & - & - 
\\
\color{black}GEN3C~\cite{ren2025gen3c}
& \color{black}16.64 	& \color{black}0.400 	    & \color{black}0.490
& \color{black}20.95 	& \color{black}0.593 	& \color{black}{0.303}
\\
\color{black}Difix3D+~\cite{wu2025difix3d}
& \color{black}17.64 	& \color{black}0.455     & \color{black}0.416 
& \color{black}\textbf{22.44} 	& \color{black}0.615 	& \color{black}\textbf{0.260} 
\\
Ours 	 
& \textbf{19.83} & \textbf{0.570} & \textbf{0.426} 
& {21.66} & \textbf{0.643} & {0.342} 
\\
\bottomrule[0.15em]
\end{tabular}
}

  \centering%
\end{table}

\subsection{Evaluation on DL3DV-Drone Datasets}
To further validate the effectiveness of our method, we conduct experiments on the DL3DV-Drone dataset. \zyqm{As shown in \Tref{table:drone}, our method consistently outperforms existing approaches on the close-up view settings and show competitive performance under the regular view, demonstrating strong robustness. Please refer to supplementary for the qualitative comparisons.}

\subsection{Ablation Study}

\noindent\textbf{The Effect of Different Estimators}
\zyqm{\Tref{table:dl3dv10k} executed an additional experiment by comparing the DUSt3R~\cite{wang2024dust3r} as the pretrained estimator against VGGT~\cite{wang2025vggt} to evaluation the effect of the fundation model. Both DUSt3R and VGGT serve as widely adopted foundation models that can be applied to most scenarios.
As shown in the table, the performance slightly degrades when using DUSt3R compared to VGGT, due to its less accurate predictions. Nevertheless, it still outperforms other methods by a large margin, particularly in the close-up setting, demonstrating the robustness of our approach.}

\begin{table}[t] \centering
    \caption{Ablation study of the proposed modules of our method, where we report the average results on the DL3DV-10K dataset.}
    \label{table:ablation}
    
\resizebox{0.49\textwidth}{!}{
\begin{tabular}{l|ccc}
\toprule[0.15em] 
Method & PSNR$\uparrow$ & SSIM$\uparrow$ & LPIPS$\downarrow$  \\ 
\midrule
ViewCrafter & 13.45 	& 0.425 	& 0.548 \\
\color{black}Baseline (Low-reso warp) & \color{black}17.07 	&\color{black}0.530 	    & \color{black}0.409   \\ 
Baseline (High-reso warp) & 17.60 	& 0.555 	& 0.413  \\ 

\phantom{00} + Hierarchical warp & 18.11 	& 0.586 	& 0.377  \\ 
\phantom{0000} + Noise suppression & 18.22 	& 0.592 	& 0.375  \\ 
\phantom{000000} + Global guidance & 18.75 	& 0.618 	& 0.369  \\ 
\phantom{00000000} + Decoder finetune & \textbf{19.66} 	& \textbf{0.643} 	& \textbf{0.346}  \\ 
\bottomrule[0.15em]
\end{tabular}
}

\end{table}

\begin{figure}[!t] \centering
    \includegraphics[width=\columnwidth]{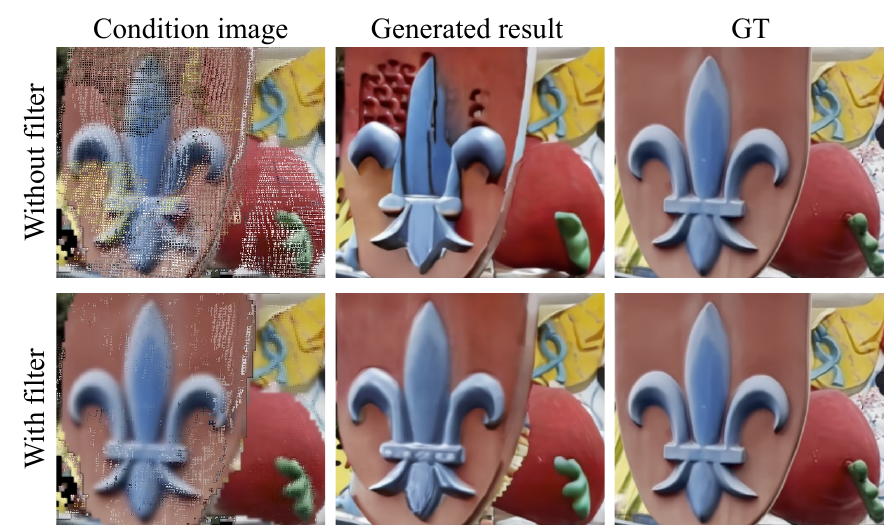}
    \caption{Effect of \occlusionfilter}
    \label{fig:occlusion}
\end{figure}

\noindent\textbf{Effect of Hierarchical Warping.}
\zyqm{To evaluate the impact of the proposed \emph{hierarchical warping} modules, we conduct ablation experiments and report quantitative results under the close-up view setting in \Tref{table:ablation}.
We first finetune the ViewCrafter with additional close-up views data, in which we conduct two "Baseline" including the low-resolution and high-resolution warping images as condition. 
Compared to the original ViewCrafter results, fine-tuning with close-up data yields clear improvements in pixel-level metrics such as PSNR and SSIM.}

\zyqm{
However, the perceptual quality remains limited, as reflected by the high LPIPS scores, suggesting that fine-grained textures and realism are still lacking. 
On the one hand, using only upsampled low-resolution warping images yields the poor performance in PSNR, as such images tend to be blurry and fail to provide high-frequency details. 
On the other hand, high-resolution images are limited by sparsity and noise. By adopting a hierarchical warping strategy, these two types of inputs complement each other, leading to consistent improvements across both pixel-level and perceptual metrics.
In particular, the significant improvement in LPIPS indicates that our hierarchical conditioning provides more reliable guidance to the diffusion model, leading to sharper and more visually appealing generations.
These results validate the effectiveness of the proposed hierarchical warping design.}

\begin{figure}[t]
  \centering
  \includegraphics[width=0.32\columnwidth]{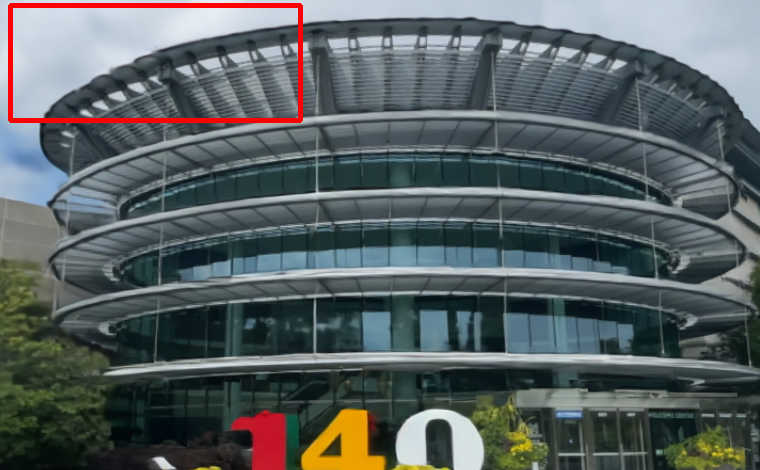}
  \includegraphics[width=0.32\columnwidth]{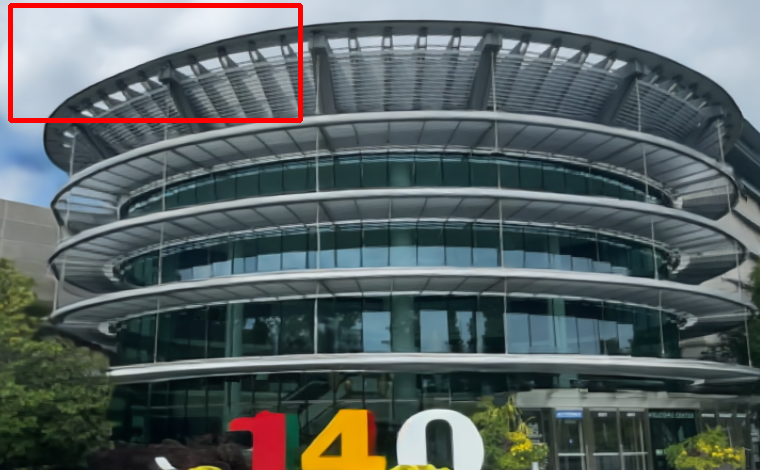}
  \includegraphics[width=0.32\columnwidth]{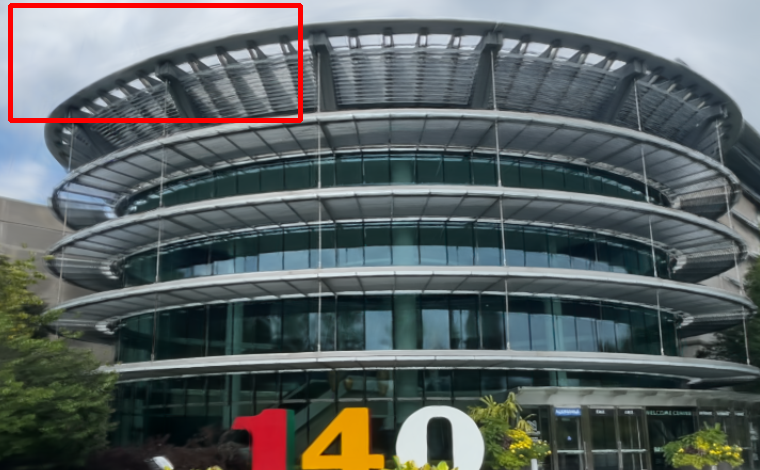}
  \\
  \includegraphics[width=0.32\columnwidth]{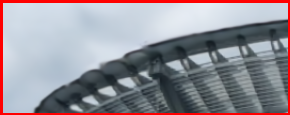}
  \includegraphics[width=0.32\columnwidth]{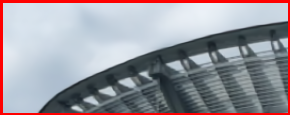}
  \includegraphics[width=0.32\columnwidth]{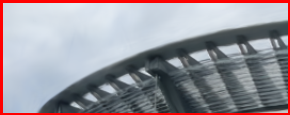}
  \\
  \includegraphics[width=0.32\columnwidth]{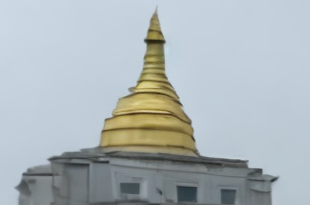}
  \includegraphics[width=0.32\columnwidth]{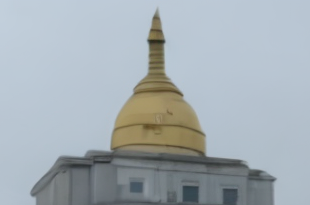}
  \includegraphics[width=0.32\columnwidth]{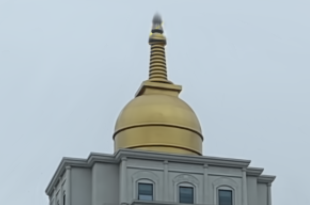}
  \\
  \includegraphics[width=0.32\columnwidth]{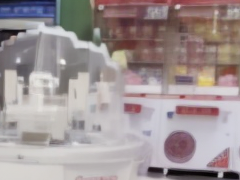}
  \includegraphics[width=0.32\columnwidth]{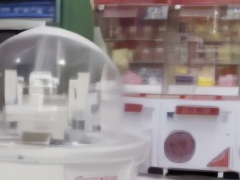}
  \includegraphics[width=0.32\columnwidth]{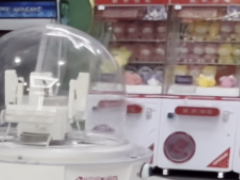}
  \\
  \vspace{-1pt}
  \makebox[0.32\columnwidth]{\scriptsize Without Global Guidance}
  \makebox[0.32\columnwidth]{\scriptsize With Global Guidance}
  \makebox[0.32\columnwidth]{\scriptsize Ground Truth}
  \\
  \caption{Effect of \globalguidance. With our global structure guidance, the geometry becomes more accurate, effectively correcting errors caused by depth inconsistencies in sparse-view settings.}
  \label{fig:global}
\end{figure}

\noindent\textbf{Effect of \Occlusionfilter.} 
To address the issue of background leakage, we introduce the \emph{\occlusionfilter} module. As shown in \Tref{table:ablation}, this module brings further improvements in quantitative performance. More importantly, it helps suppress artifacts caused by background leakage. 
\Fref{fig:occlusion} presents an example where background points are visible in the conditioning image, resulting in noticeable artifacts in the generated output. 
By applying our \occlusionfilter strategy, these background leakages are effectively suppressed, leading to plausible and high-fidelity generations. \zyqm{Please refer to supplementary material for more illustration.}

\noindent\textbf{Effect of \GlobalGuidance.}
\Fref{fig:global} shows an example where inaccurate estimated depths lead to errors in the conditioning images, \eg, duplicated tower tips appear. Incorporating the proposed \emph{\globalguidance} can correct such artifacts and hence improve geometric structure, demonstrating the effectiveness of our design.
We can see frome \Tref{table:ablation} that with global guidance, the performance can be improved by 0.5 db of PSNR. 
Moreover, finetuning the VAE decoder can further improve the perfromance of our model, as reported in \Tref{table:ablation}.

\noindent\textbf{Analysis of 3DGS Optimization.} 
To evaluate the impact of 3DGS optimization, we conduct a comparison between the diffusion baseline and various 3DGS enhancement strategies in Table \ref{table:3dgs}, which reports the average metrics on the DL3DV-10K and DL3DV-Drone dataset. We can see from the table that the vallina 3DGS configuration leads to noticeable degradation in both PSNR and LPIPS metrics, which primarily stems from the inherent inconsistency of diffusion-based generation. To address this limitation, we introduce confidence-based optimizations in terms of image-level and pixel-level, which enhance the performance of 3D reconstruction over the diffusion method.

\begin{table}[t]
    \begin{center}
    \caption{Analysis of 3DGS optimization with image-level confidence and pixel-level confidence. We report the average metrics on the DL3DV-10K and DL3DV-Drone dataset.}
    \label{table:3dgs}
    
\resizebox{0.49\textwidth}{!}{
\begin{tabular}{lccc|ccc}
\toprule[0.15em]
\multirow{2}{*}{Method} & \multicolumn{3}{c|}{Close-up View} & \multicolumn{3}{c}{Regular View} \\
\cmidrule(lr){2-4} \cmidrule(lr){5-7}
& PSNR$\uparrow$ & SSIM$\uparrow$ & LPIPS$\downarrow$ 
& PSNR$\uparrow$ & SSIM$\uparrow$ & LPIPS$\downarrow$ \\
\midrule
Ours with diffusion
& 19.61 	& 0.608 	& \textbf{0.358} 	& 19.93 	& 0.602 	& \textbf{0.360}
\\
Ours with vallina 3DGS 
& 19.40 	& 0.625 	& 0.403 	& 19.99 	& 0.642 	& 0.386 
\\
\phantom{00} + Image confidence
& 19.64 	& 0.633 	& 0.397 	& 20.24 	& 0.653 	& 0.378 
\\
\phantom{0000} + Pixel confidence
& \textbf{19.80} 	& \textbf{0.637} 	& 0.391 	& \textbf{20.33} 	& \textbf{0.656} 	& 0.371 
\\
\bottomrule[0.15em]
\end{tabular}
}

    \end{center}
\end{table}

\subsection{Evaluation on Cross Dataset}
\zyqm{To evaluate the generalization of our method, we compare our method against DepthSplat~\cite{xu2024depthsplat} and ViewCrafter~\cite{yu2024viewcrafter}} on two additional unseen datasets, RealEstate10K~\cite{zhou2018stereo} and ACID~\cite{liu2021infinite} datasets. The quantitative results are shown on \Tref{table:crossdataset}, where our method exhibits strong generalization with higher metrics in terms of PSNR, SSIM, and LPIPS. Moreover, \fref{fig:crossdataset} presents the visualization results on the RealEstate10K, demonstrating the generalization ability of our method.

\begin{table}[!t] \centering
    \captionof{table}{Cross-dataset generalization. Our method exhibits strong generalization ability on unseen datasets such as RealEstate10K and ACID, despite not being trained on them.}
    \label{table:crossdataset}
    \resizebox{0.49\textwidth}{!}{
\begin{tabular}{lccc|ccc}
\toprule[0.15em]
\multirow{2}{*}{Method} & \multicolumn{3}{c|}{RealEstate10K} & \multicolumn{3}{c}{ACID} \\
\cmidrule(lr){2-4} \cmidrule(lr){5-7}
& PSNR$\uparrow$ & SSIM$\uparrow$ & LPIPS$\downarrow$ 
& PSNR$\uparrow$ & SSIM$\uparrow$ & LPIPS$\downarrow$ \\
\midrule
\color{black}ViewCrafter
& \color{black}16.27 	& \color{black}0.594 	& \color{black}0.501 
& \color{black}16.52 	& \color{black}0.590 	& \color{black}0.525 	
\\
DepthSplat
& 19.18 	& 0.552 	& 0.334	
& 20.29 	& 0.675 	& 0.338  	 
\\
Ours 	 
& \textbf{22.44} & \textbf{0.716} & \textbf{0.297} & \textbf{23.79} & \textbf{0.774} & \textbf{0.281} 
\\
\bottomrule[0.15em]
\end{tabular}
}

    \vspace{6pt}
    \includegraphics[width=\columnwidth]{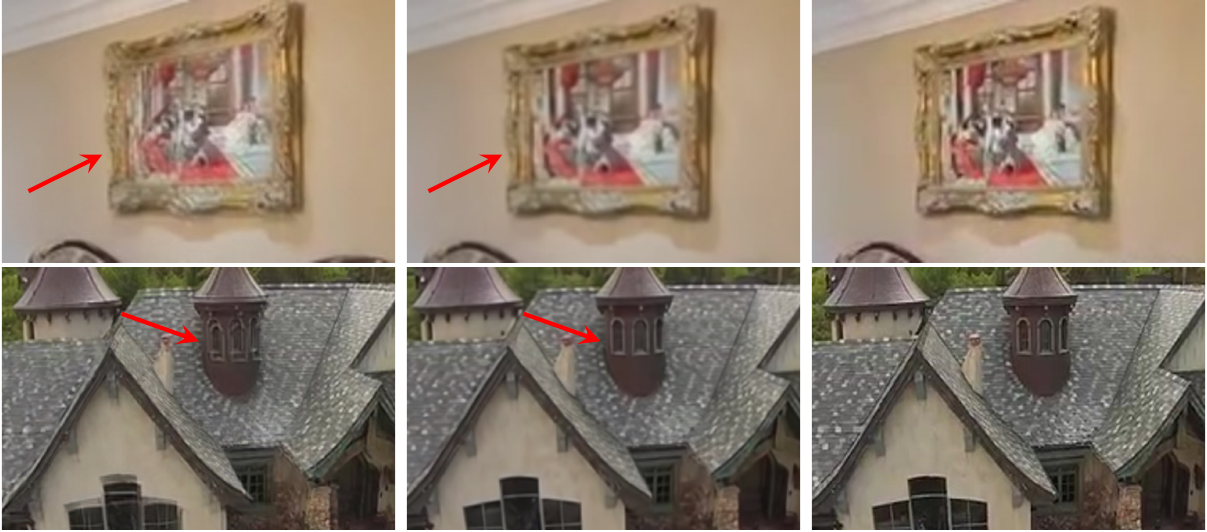}
    \makebox[0.32\columnwidth]{\scriptsize DepthSplat}
    \makebox[0.32\columnwidth]{\scriptsize Ours}
    \makebox[0.32\columnwidth]{\scriptsize Ground Truth}
    \caption{Cross dataset generalization from DL3DV to RealEstate10K.}
    \label{fig:crossdataset}
    \vspace{-1.2em}
\end{table}

\section{Conclusion}
\label{sec:Conclusion}

In this paper, we have introduced a method, \MethodName, for close-up novel view synthesis and 3D reconstruction from sparse input views based on point-conditioned video diffusion model.
Our method enhances the conditioning quality for video diffusion models by proposing a hierarchical warping strategy and an \occlusionfilter  module, effectively addressing the challenges of sparsity and background leakage under close-up settings.
To further ensure geometric consistency, we incorporate a \globalguidance mechanism derived from multi-view consistency checks, which improves the fidelity and coherence of generated views.
Finally, the generated views are used to supervise 3D Gaussian Splatting for photorealistic and detail-preserving 3D reconstruction.
Extensive experiments on challenging datasets demonstrate the effectiveness of our method, particularly in scenarios requiring close-up fine-grained inspection.

\noindent\textbf{Discussion.}
Our method relies on a pretrained estimator to predict camera poses and depth maps. When the estimator fails, the resulting errors in depth or pose can degrade the quality of both conditioning inputs and final reconstruction.
Moreover, our method involves one extra DDIM inference runtime compared to ViewCrafter, as required by the global geometric guidance, though DepthFusion itself remains efficient (within 2 seconds).
Finally, since our approach is based on multi-step diffusion sampling, extending it to a single-step sampling scheme~\cite{wu2025difix3d} for improved efficiency remains an interesting direction for future work.

\bibliographystyle{IEEEtran}
\bibliography{ref}

\vfill

\end{document}